\documentclass[lettersize,journal]{IEEEtran}
\usepackage{amsmath,amsfonts}
\usepackage{algorithmic}
\usepackage{algorithm}
\usepackage{array}
\usepackage[caption=false,font=footnotesize,labelfont=rm,textfont=rm]{subfig}
\usepackage{textcomp}
\usepackage{stfloats}
\usepackage{url}
\usepackage{verbatim}
\usepackage{graphicx}
\usepackage{cite}
\usepackage{multirow}
\usepackage{multicol}
\usepackage{arydshln}
\usepackage{booktabs}
\usepackage{amsthm,amsmath,amssymb}
\usepackage{mathrsfs}
\usepackage{color}
\begin{document}

\title{Group Multi-View Transformer for 3D Shape Analysis with Spatial Encoding}

\author{Lixiang Xu, \textit{Member, IEEE}, Qingzhe Cui, Richang Hong, \textit{Senior Member, IEEE}, Wei Xu,
Enhong Chen, \textit{Senior Member, IEEE}, Xin Yuan, \textit{Member, IEEE}, Chenglong Li and Yuanyan Tang, \textit{Life Fellow, IEEE}\thanks{This work was financially supported by National Natural Science Foundation of China (62176085, 62172458), Scientific Research Innovation Team in Colleges and Universities of Anhui Province (2022AH010095) and Industry-University-Research Cooperation Project (GP/026/2020 and HF-010-2021) Zhuhai City, Guangdong Province, China. \textit{(Corresponding author: Richang Hong and Enhong Chen. Equal Contribution: Lixiang Xu and Qingzhe Cui.)}

Lixiang Xu, Qingzhe Cui and Wei Xu are with the College of Artificial Intelligence and Big Data, Hefei University, Hefei 230027, China (e-mail: xulixianghf@163.com; cuiqz886@163.com; xuw981019@gmail.com).

Richang Hong is with the School of Computer Science and Information Engineering, Hefei University of Technology, Hefei 230009, China (e-mail: hongrc@hfut.edu.cn).


Enhong Chen is with the Anhui Province Key Laboratory of Big Data Analysis and Application, School of Computer Science and Technology, University of Science and Technology of China, Hefei, Anhui 230000, China (e-mail: cheneh@ustc.edu.cn).


Xin Yuan is with the School of Electrical and Mechanical Engineering, The University of Adelaide, Adelaide, SA 5005, Australia (e-mail:
xin.yuan@adelaide.edu.au).

Chenglong Li is with the School of Artificial Intelligence, Anhui University, Hefei, 230601, China (lcl1314@foxmail.com).

Yuanyan Tang is with the Zhuhai UM Science and Technology Research Institute, FST University of Macau, Macau (e-mail: yytang@umac.mo).
}
}
\markboth{Journal of \LaTeX\ Class Files,~Vol.~14, No.~8, August~2021}%
{Shell \MakeLowercase{\textit{et al.}}: A Sample Article Using IEEEtran.cls for IEEE Journals}


\maketitle

\begin{abstract}
In recent years, the results of view-based 3D shape recognition methods have saturated, and models with excellent performance cannot be deployed on memory-limited devices due to their huge size of parameters. To address this problem, we introduce a compression method based on knowledge distillation for this field, which largely reduces the number of parameters while preserving model performance as much as possible. Specifically, to enhance the capabilities of smaller models, we design a high-performing large model called Group Multi-view Vision Transformer (GMViT). In GMViT, the view-level ViT first establishes relationships between view-level features. Additionally, to capture deeper features, we employ the grouping module to enhance view-level features into group-level features. Finally, the group-level ViT aggregates group-level features into complete, well-formed 3D shape descriptors. Notably, in both ViTs, we introduce spatial encoding of camera coordinates as innovative position embeddings. Furthermore, we propose two compressed versions based on GMViT, namely GMViT-simple and GMViT-mini. To enhance the training effectiveness of the small models, we introduce a knowledge distillation method throughout the GMViT process, where the key outputs of each GMViT component serve as distillation targets. Extensive experiments demonstrate the efficacy of the proposed method. The large model GMViT achieves excellent 3D classification and retrieval results on the benchmark datasets ModelNet, ShapeNetCore55, and MCB. The smaller models, GMViT-simple and GMViT-mini, reduce the parameter size by 8 and 17.6 times, respectively, and improve shape recognition speed by 1.5 times on average, while preserving at least 90\% of the classification and retrieval performance. The code is available at https://github.com/bigdata-graph/GMViT.
\end{abstract}


\section{Introduction}
\IEEEPARstart{W}{ith} the popularity of various 3D acquisition devices, the volume of 3D data has surged, which in turn has facilitated a shift from theoretical research on 3D data to experimental research based on deep learning. The main deep learning methods about 3D shape analysis are voxel-based methods \cite{maturana2015voxnet, wu20153d, brock2016generative}, point-based methods \cite{wang2019dynamic,zhou2021adaptive,xu2018spidercnn,li2018pointcnn,qi2017pointnet,qi2017pointnet++,guo2021pct, ma2022rethinking, zeid2023point2vec, srivastava2021exploiting} and view-based methods \cite{han20193d2seqviews, ma2018learning, feng2018gvcnn, han2018seqviews2seqlabels, kanezaki2018rotationnet, wei2020view, xu2021multi, yu2018multi, yang2019learning, lin2022multi}. All of the above methods have been widely applied in various fields such as autonomous driving, virtual/augmented reality, and medical diagnosis.

Voxel-based methods extend 2D pixels to 3D space and extract their features by convolutional neural networks (CNNs) equipped with 3D convolutional kernel. Although this type of approach can achieve satisfactory performance, the memory footprint and computational consumption caused by increasing voxel resolution are significant. Point-based methods generate point clouds by scanning the surface of 3D objects with devices such as LiDAR, then learn geometric features on the surface of the point clouds through deep learning methods, and finally aggregate the extracted local information into global features utilizing symmetry functions. The view-based methods render the 3D target from different angles to get multiple views, then extract the information from individual views separately, and finally aggregate all the view features into 3D shape descriptors. 

How to efficiently fuse multiple view features and avoid redundancy of features has always been the most important issue for this class of methods. This is because seeing an object from only one angle is partial and the views rendered from adjacent angles have a high degree of similarity. To solve the above problem, a number of view feature fusion methods \cite{su2015multi, feng2018gvcnn, han2018seqviews2seqlabels, nie2021dan, wei2020view} have been proposed. Initially, using the symmetry of pooling functions is the most direct means to aggregate multiple view features into a 3D shape descriptor, but such simple pooling operations ignore the complementary relationships between views, which inevitably leads to loss of information. Thus, a number of approaches attempting to fully fuse multiple view features have since been introduced, such as using group pooling to capture the relationships of similar views \cite{feng2018gvcnn}, treating multiple views as a set of ordered sequences and capturing the sequential relationship between them via recurrent neural networks \cite{graves2012long} (RNN) \cite{han2018seqviews2seqlabels}, trying to learn the optimal rendering positions of the camera to obtain more expressive images \cite{hamdi2021mvtn}, employing the self-attention mechanism of Vision Transformer \cite{dosovitskiy2020image} (ViT) to obtain global information between views \cite{nie2021dan}, and considering the spatial structure of the views as a graph and utilizing a graph convolutional neural network \cite{xu2021multiple, xu2021deep} (GCN) to aggregate information between views \cite{wei2020view}. 

Despite the advancements made by the aforementioned methods in addressing the issue of view feature fusion, some limitations still persist. For instance, the group pooling approach \cite{feng2018gvcnn} incorporates group feature pooling before global pooling, yet this intermediary step merely reduces the pooling scale, resulting in some information loss. To compensate for the inevitable loss, it becomes crucial to allow all features to interact fully before pooling. Consequently, our study introduces a novel approach that establishes relationships between view-level features and group-level features before applying group and global pooling independently.
Furthermore, the RNN-based methods \cite{han2018seqviews2seqlabels, xu2019deeply, jin2021prema, ma2018learning} primarily consider 1D sequential relationships among views, while the self-attention-based approach \cite{nie2021dan, chen2021mvt, lin2023multi} uses traditional position embeddings to establish view relationships, inadvertently overlooking the spatial relationships among views. Given that multi-views are generated by placing the camera at various coordinates around the 3D object, which inherently carry vital positional information, we propose to map the rendering coordinates of the views to potential position embeddings when establishing view relationships through ViT.

Additionally, various 3D shape recognition methods \cite{wei2020view, kanezaki2018rotationnet, xu2021multi, chen2021mvt} have demonstrated exceptional performance, reaching a saturation point on certain 3D shape recognition datasets. Despite their improved performance, these methods tend to increase model parameters and reduce computation speed, restricting deployment to high-capability machines and limiting their application on mobile devices. Thus, it becomes necessary to compress the models while maintaining their excellent performance. Recent research has focused on knowledge distillation (KD) methods \cite{hinton2015distilling, yang2022vitkd, touvron2021training, zhang2022minivit} for model compression. The concept was initially introduced by Hinton et al. \cite{hinton2015distilling} and has since evolved, with KD involving the use of a high-performance teacher network's output as soft labels for a low-performance student network. While most KD advancements were designed for CNN models, several KD methods \cite{yang2022vitkd, touvron2021training, zhang2022minivit} tailored for the ViT model have recently emerged, demonstrating their efficacy in feature or class token distillation through extensive experimentation.

While extensive research has focused on KD in the field of 2D image recognition, its application in 3D shape recognition remains unexplored. 3D data comprises complex but more comprehensive information compared to 2D data, necessitating additional computational steps for effective information extraction. For instance, in 2D domain, the network model only needs to extract information from a single image to recognize an object. However, in the 3D multi-view domain, the network model must process individual images and integrate valuable information from multiple images while discarding redundant information. Therefore, it is necessary to compress the multi-view processing model. 

In multi-view knowledge distillation, the choice of intermediate outputs from the teacher model as distillation targets should consider several factors. First, selecting outputs from structurally complex modules like the self-attention mechanism in ViT can transfer more sophisticated feature information that is difficult for the weaker student model to learn. Second, outputs from information-rich modules like fully-connected layers contain more global features and can also be beneficial distillation targets. Additionally, combining outputs from different abstraction levels, both low-level and high-level semantics, can enable more comprehensive feature distillation. Analyzing each module's impact on the downstream task and selecting influential outputs is another strategy. Overall, choosing intermediate outputs with high information content and significance to guide the student model in learning the teacher’s core knowledge enables effective distillation. Specifically, this paper performs feature distillation from the CNN, view-level ViT, and group-level ViT modules to transfer multi-scale information. The group tokens are also distilled to align grouping. Logit distillation further provides holistic guidance. This multifaceted approach allows comprehensive knowledge transfer from teacher to student.

The main contributions of this paper are as follows: 
\begin{itemize}
    \item Proposing the Group Multi-view Vision Transformer (GMViT), a 3D shape recognition model that utilizes the rendering coordinates of views as position embeddings for the first time. 
    This approach achieves state-of-the-art classification and retrieval results on benchmark datasets.
    
    \item Designing compressed versions of GMViT, namely GMViT-simple and GMViT-mini, which significantly reduce the size of model parameters and computational complexity while improving the speed of 3D object recognition.
    
    \item Pioneering the application of the knowledge distillation method in the field of 3D shape recognition. GMViT serves as the teacher model, while GMViT-simple and GMViT-mini are utilized as student models. The student models preserve the majority of the teacher model's performance through feature-based, group token-based, and logit-based distillation methods.

\end{itemize}

The rest of the paper is organized as follows. Section \ref{sec2} presents the related work. Section \ref{sec3} details the proposed method. Section \ref{sec4} presents the experimental results and analysis. Section \ref{sec5} summarizes the full paper.

\section{Related Work}
\label{sec2}
This section provides a review of voxel-based, point-based and view-based 3D shape analysis methods. In addition, existing works on knowledge distillation are also reviewed.
\subsection{Voxel-Based Methods}
The voxel-based methods divide the 3D space into voxel units and construct a shape representation of the 3D object on them. The initial volume processing method is 3D Shapenet \cite{wu20153d}, where the probability distribution of binary variables on a 3D voxel grid is obtained by learning a convolutional deep belief network. VoxNet \cite{maturana2015voxnet} utilizes CNNs equipped with 3D convolution to output voxel occupancy on meshes. 3D convolution has higher complexity than 2D convolution, which leads to an exponential increase in time complexity and computational cost of such methods when the depth of the network or the resolution of the voxels increases. Therefore, some methods featuring low consumption and high efficiency have been proposed. O-CNN \cite{wang2017cnn} is a CNN-based octree, aiming to use octrees to divide 3D shapes at different scales using octrees, which greatly improves the efficiency of voxel processing.

\subsection{Point-Based Methods}
Point clouds, compared to other modalities, have a simple representation comprising the coordinates of points on a 3D shape's surface. PointNet \cite{qi2017pointnet} processes point clouds directly using deep networks, extracting features with MLP and obtaining global features through pooling, effectively addressing permutation invariance and disorder. PointNet++ \cite{qi2017pointnet++} improves segmentation by incorporating neighborhood information, overcoming PointNet's limitation. Wang et al. \cite{wang2019dynamic} proposed the EdgeConv module, establishing edges between points and neighbors using KNN. Lin et al. \cite{lin2020convolution} used a deformable kernel with a 3D graph convolutional neural network. AdaptConv \cite{zhou2021adaptive} developed an adaptive kernel considering central points and neighbors. With the success of the self-attention mechanism, subsequent models like PCT \cite{guo2021pct} and Point Transformer \cite{zhao2021point} aim to establish global relationships among all points.

\subsection{View-Based Methods}
View-based methods represent 3D objects through a set of 2D views rendered at different angles. MVCNN \cite{su2015multi}, the earliest study of this kind of method on deep learning, uses a set of CNNs with shared weights to extract features of all views, and then feeds these features to a pooling function to obtain shape descriptors. Although the process is simple, it provides a very valuable reference for subsequent studies. GVCNN \cite{feng2018gvcnn} incorporated a hierarchical structure that divides similar view features into groups and applies pooling functions within each group and layer. This approach aims to mitigate feature loss resulting from direct employment of global pooling. In contrast to GVCNN, we introduce the Vision Transformer before group pooling and global pooling stages. This approach facilitates the establishment of global relationships between view-level features and group-level features, respectively. Consequently, it effectively mitigates information loss resulting from pooling. Wei et al. \cite{wei2020view} considered a set of views as a graph, aggregate the neighboring features of each view node through GCN, and aggregate view features at different scales using a hierarchical structure. The MVTN \cite{hamdi2021mvtn}  proposed by Hamdi et al. improves the representation of 3D objects by learning the optimal rendering positions of the views.

Some methods utilize the order of view arrangement to enhance the learning of shape descriptors. These methods organize a set of views into a specific sequence based on predefined rules and subsequently utilize RNNs to capture temporal features among the views. Ma et al. \cite{ma2018learning} assigned weights and aggregated view features from each time step of the Long Short-Term Memory network \cite{graves2012long} (LSTM) to derive global features. Xu et al. \cite{xu2019deeply} captured the bi-directional dependency of view sequences by employing a Bi-directional Long Short-Term Memory network \cite{huang2015bidirectional} (Bi-LSTM). Jin et al. \cite{jin2021prema} introduced a partial-based recurrent feature aggregation module, which utilizes LSTM to accumulate features from specific regions within each view over time. The SeqViews2SeqLabels \cite{han2018seqviews2seqlabels} model primarily comprises an Encoder RNN and a Decoder RNN. The Encoder RNN is responsible for aggregating global features from a sequence of views, while the Decoder RNN is utilized for predicting the label of a 3D shape. In contrast, the 3D2SeqViews \cite{han20193d2seqviews} model does not rely on an RNN structure to acquire sequence features. Instead, it employs hierarchical attention modules to aggregate view features into global features.

Additionally, there exist methods that leverage the self-attention mechanism of ViT to capture the global relationships among views. Chen et al. proposed MVT \cite{chen2021mvt}, a method that initially employs a Local Transformer Encoder to capture relationships between patches within each view individually. Subsequently, a Global Transformer Encoder is utilized to enable comprehensive interaction among patches from all views. MVDAN \cite{wang2022multi} combines the two features produced by the view space attention block and the channel attention block to generate compact shape descriptors. Nie et al. \cite{nie2021dan} broke the conventional multi-head self-attention approach and facilitated the fusion of multi-view features through the utilization of stacked deep self-attention. Lin et al. \cite{lin2023multi} highlighted that aggregating neighboring views could result in feature redundancy. Therefore, they introduced Mid-Range and Long-Range views to complement the Short-Range view features. This approach involved aggregating view features at each scale using the ViT Encoder. The aforementioned methods employ regular position embeddings, such as \cite{dosovitskiy2020image}, during the aggregation of view-level features using ViT. Views are generated by cameras that are discretely positioned in 3D space, and unlike patches of 2D images, they do not exhibit fixed front-to-back dependencies. Consequently, our GMViT, maps the rendering coordinates of each view to novel position embeddings.


\subsection{Knowledge Distillation}
KD, a highly effective method for enhancing the performance of small models, has generated significant attention in recent years. Hinton et al. \cite{hinton2015distilling} pioneered the usage of soft labels derived from the teacher model's output to enhance the training of the student model. This approach not only significantly compressed the small model but also yielded remarkable performance improvements. Initially, KD was predominantly employed for compressing CNN-based models. However, Touvron et al. \cite{touvron2021training} extended the application of KD to ViT-based models and demonstrated its viability. The recently proposed miniViT \cite{zhang2022minivit} by Zhang et al. employs self-attention distillation and Hidden-State distillation, which is feature-based distillation. Yang et al. \cite{yang2022vitkd} propose a novel approach for feature-based ViT distillation, which utilizes a special method to distill three distinct components of the teacher model. All the aforementioned methods are utilized for 2D image recognition, while the performance of traditional 3D shape recognition methods based on feature aggregation has reached a saturated point in recent years. Therefore, this paper aims to introduce knowledge distillation into the domain of multi-view recognition for the first time.

\begin{figure*}[t]
	\centering
	\includegraphics[width=0.8\linewidth]{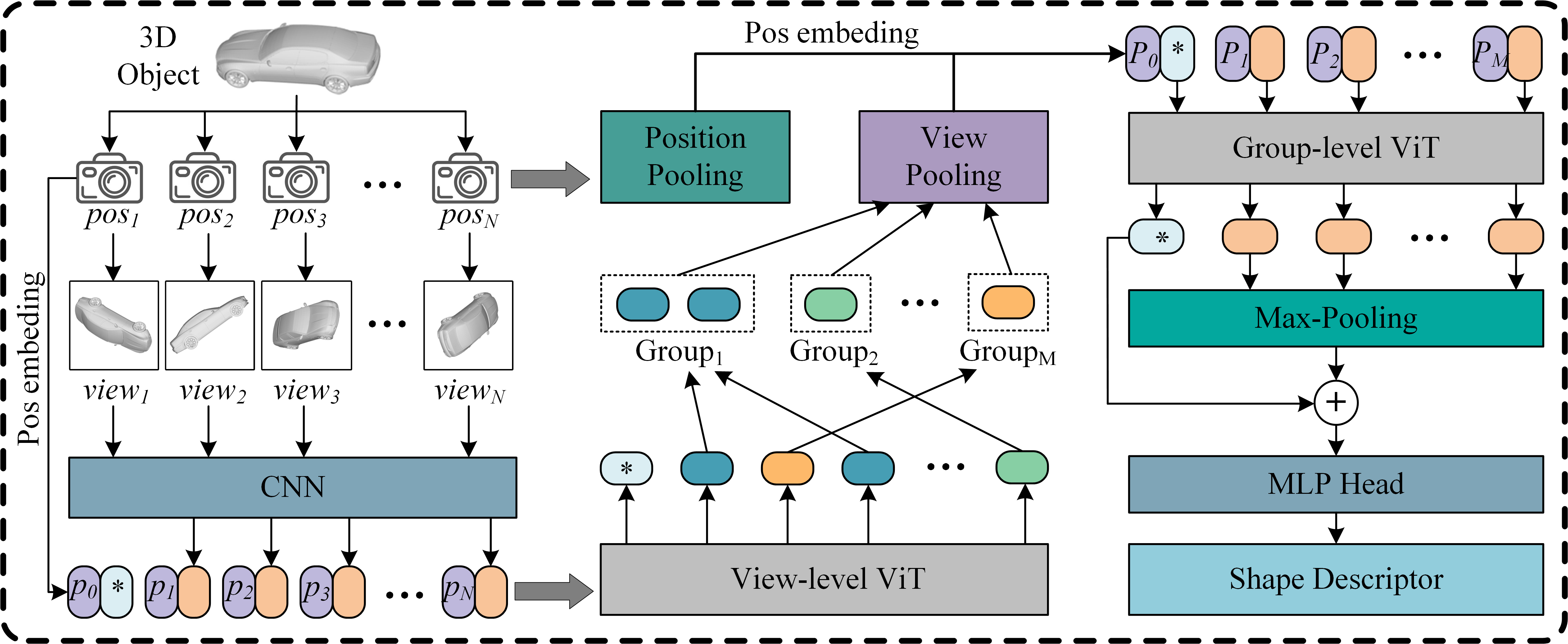}
	\caption{The general framework diagram of Group Multi-view Vision Transformer.}
	\label{fig1}
\end{figure*}
\section{Proposed method}
\label{sec3}
\subsection{Group Multi-view Vision Transformer}
\subsubsection{Overview}
The overall framework of GMViT is shown in Fig. \ref{fig1}. Initially, we utilize $N$ cameras positioned at location $pos=\{{pos}_1,\ {pos}_2,\ \ldots,\ {pos}_N\}\in \mathbb{R}^{N\times3}$ to render the 3D objects, generating a corresponding set of views, $VIEW=\{{view}_1,\ {view}_2,\ \ldots,\ {view}_N\}$. Then, we employ a set of CNNs with shared weights to extract the features $F_v=\{f_1,\ f_2,\ \ldots,\ f_N\}\in \mathbb{R}^{N\times D}$ from all the views. Subsequently, the position information is embedded into the view feature $F_v$ with class token and fed into the view-level ViT. Within the view-level ViT, the position embeddings of the views are derived based on their respective camera positions, $pos$. Next, we dynamically group and pool the view features obtained from the view-level ViT along with the $pos$. Lastly, the view features of each group are sequentially aggregated to generate the final 3D shape descriptor. This aggregation process involves the group-level ViT, Max-Pooling, and MLP Head.

\subsubsection{View-level ViT}
Before inputting the CNN-extracted view features $F_v$ into the ViT, it is necessary to perform a position embedding of these features and the class token $f_{cls}$. In contrast to existing multi-view approaches that employ ViT, we introduce a novel position embedding method. This method utilizes a MLP to map the camera positions $pos$ of the captured views to the position embeddings $p_v=\{p_1,\ p_2,\ ...,\ p_N\}\in \mathbb{R}^{N\times D}$ of the view features: 
\begin{equation}
	p_v=mlp(pos)
	\label{eqa1}
\end{equation}
where $mlp$ stands for MLP. Then the process of embedding position information for the view features is:
\begin{equation}
	p_V=\left[p_{cls},\ p_1,\ p_2,\ ...,\ p_N\right] \in \mathbb{R}^{(N+1)\times D}		
	\label{eqa2}
\end{equation}
\begin{equation}
	F_V=\left[f_{cls},\ f_1,\ f_2,\ \ldots,\ f_N\right] \in \mathbb{R}^{(N+1)\times D}
	\label{eqa3}
\end{equation}
\begin{equation}
	F_V^\ast=p_V+F_V
	\label{eqa4}
\end{equation}
where $\left[ \cdot,\ \cdot \right]$ denotes the concatenation operation, $F_V^\ast$ represents the input feature of view-level ViT, and the class token $f_{cls}$ along with its corresponding position embedding $p_{cls}$ are acquired through a learning process. The position information from the cameras, distributed in 3D space, is incorporated into the view features, thereby enhancing the spatial information in the 3D shape descriptors. Subsequently, the features $F_V^\ast$ are inputted into the view-level ViT, resulting in the generation of interacted features $F_{{ViT}_V}=\{f_{{ViT}_{cls}},\ f_{{ViT}_{1}},\ f_{{ViT}_{2}}\ \ldots,\ f_{{ViT}_{N}}\}\in \mathbb{R}^{(N+1)\times D}$. The view-level ViT, denoted as ${ViT}_V=\{{ViT}_{v_1},\ {ViT}_{v_2},\ \ldots,\ {ViT}_{v_L}\}$, comprises a series of $L$-layer ViTs.
The process is:
\begin{equation}
    F_{{ViT}_V}={ViT}_{v_L}(\ldots({ViT}_{v_1}(F_V^\ast)))
    \label{eqa5}
\end{equation}

\subsubsection{View grouping}
To obtain 3D information at different scales, inspired by \cite{feng2018gvcnn}, we group the view features $F_{{ViT}_{view}}=\{f_{{ViT}_{1}},\ f_{{ViT}_{2}}\ \ldots,\ f_{{ViT}_{N}}\}$ of the view-level ViT output. First, we define a feature set $G_F=\ \{G_{F_1},\ G_{F_2},\ ...,\ G_{F_M}\}$ and a position set $G_P=\ \{G_{P_1},\ G_{P_2},\ ...,\ G_{P_M}\}$. Subsequently, we utilize an MLP along with a sigmoid activation function to map the view features $F_{{ViT}_{view}}$ to the group token set $Token=\{t_1,\ t_2,\ ...,\ t_i,\  ...,\ t_M\}\in \mathbb{R}^{M\times1}$:
\begin{equation}
    Token=sigmoid(mlp(F_{{ViT}_{view}}))
    \label{eqa6}
\end{equation}
If the $i$-th view's group token satisfies:
\begin{equation}
    (m-1)/M\le t_i<m/M
    \label{eqa7}
\end{equation}
then the feature $F_{{ViT}_{i}}$ corresponding to the $i$-th view, along with the position $p_i$ of its camera, is assigned to the $m$-th feature group $G_{F_m}$ and the position group $G_{P_m}$, respectively $(1\le m\le M,\ m\in Z)$. Ultimately, the feature information and position information of the views will be fused independently within their respective groups. The group-level view features $F_g=\{F_1,\ F_2,\ ...,\ F_M\}\in \mathbb{R}^{M\times D}$ are acquired by employing the maximum pooling function for aggregation, as follows: 
\begin{equation}
    F_g=\{max(G_{F_1}),\ max(G_{F_2}),\ ...,\ max(G_{F_M})\}
    \label{eqa8}
\end{equation}
where $max$ denotes maximum pooling. Regarding the position coordinates within each group, we compute their center-of-mass positions, which serve as the updated position information ${POS}_G=\{{POS}_1,\ {POS}_2,\ ...,\ {POS}_m,\ ...,\ {POS}_M\}$. Suppose that $G_{P_m}=\{(x_1,\ y_1,\ z_1),\ (x_2,\ y_2,\ z_2),\ ...,\ (x_u,\ y_u,\ z_u)\}$ represents the position set of the $m$-th group. In this case, the computation of ${POS}_m=(x_m,\ y_m,\ z_m)\in \mathbb{R}^3$ is performed as follows: 
\begin{equation}
    \left\{
             \begin{array}{lr}
             x_m=(x_1+x_2+...+x_u)/u, & 
             \\
             y_m=(y_1+y_2+...+y_u)/u, & 
             \\
             z_m=(z_1+z_2+...+z_u)/u.
             \end{array}
    \right.
    \label{eqa9}
\end{equation}

\subsubsection{Group-level ViT}
The group-level ViT ${VIT}_G=\{{VIT}_{g_1},\ {VIT}_{g_2},\ ...,\ {VIT}_{g_K}\}$ consists of $K$ layers of ViT arranged in series, similar to the view-level ViT. Likewise, the processing steps for the group-level feature $F_g$ using ${VIT}_G$ follow a similar pattern to those for the view-level feature $F_v$ utilizing the view-level ViT. First, the group-level position information ${POS}_G$ is embedded into the group-level feature $F_g$: 
\begin{equation}
	P_g=mlp({POS}_G)=\{P_1,\ P_2,\ ...,\ P_M\}\in \mathbb{R}^{M\times D}
	\label{eqa10}
\end{equation}
\begin{equation}
	P_G=\left[P_{cls},\ P_1,\ P_2,\ ...,\ P_M\right]\in \mathbb{R}^{(M+1)\times D}
	\label{eqa11}
\end{equation}
\begin{equation}
	F_G=\left[F_{cls},\ F_1,\ F_2,\ \ldots,\ F_M\right]\in \mathbb{R}^{(M+1)\times D}
	\label{eqa12}
\end{equation}
\begin{equation}
	F_G^\ast=P_G+F_G
	\label{eqa13}
\end{equation}
where $F_G^\ast$ denotes the input feature of group-level ViT, and the class token $F_{cls}$ along with its corresponding position embedding $P_{cls}$ are learnable.
$F_{{VIT}_G}=\ \{F_{{VIT}_{cls}},\ F_{{VIT}_1},\ F_{{VIT}_2},\ \ldots,\ F_{{VIT}_M}\}\in \mathbb{R}^{(M+1)\times D}$ is obtained by utilizing the group-level ViT with $F_G^\ast\in \mathbb{R}^{(M+1)\times D}$ as the input feature: 
\begin{equation}
    F_{{VIT}_G}={VIT}_{g_L}(...({VIT}_{g_1}(F_G^\ast)))
    \label{eqa14}
\end{equation}
Subsequently, we concatenate the maximum pooled group features $F_{{VIT}_{group}}=\{F_{{VIT}_1},\ F_{{VIT}_2},\ \ldots,\ F_{{VIT}_M}\}\in \mathbb{R}^{M\times D}$ with the class token $F_{{VIT}_{cls}}\in \mathbb{R}^D$, and input this concatenated representation into the MLP Head to generate the final 3D shape descriptor $F_D\in \mathbb{R}^D$: 
\begin{equation}
	F_D=mlp(F_{{VIT}_{cls}},\ max(F_{{VIT}_{group}}))
	\label{eqa15}
\end{equation}

\subsubsection{Feature Classification}
Once the shape descriptor $F_D$ is obtained, it is utilized for downstream tasks. In order to obtain the prediction result $F_{pred}$ of the model, we introduce multiple MLPs to reduce the dimensionality of the feature $F_D$. Additionally, between each pair of MLPs, we include BatchNorm1d and ReLU activation functions to expedite the convergence of model:
\begin{equation}
	F_D^1=ReLU(Norm(mlp(F_D)))
	\label{eqa16}
\end{equation}
\begin{equation}
	F_D^2=ReLU(Norm(mlp(F_D^1)))
	\label{eqa17}
\end{equation}
\begin{equation}
	F_{pred}=mlp(F_D^2)
	\label{eqa18}
\end{equation}
where $Norm$ denotes BatchNorm1d. The entire network is optimized by minimizing the cross-entropy loss between the prediction result $F_{pred}$ and the Ground Truth.

\subsection{GMViT-simple and GMViT-mini}
In this section we introduce two lightweight variants of GMViT, namely GMViT-simple and GMViT-mini. 
Recent 3D shape recognition methods typically leverage pre-trained CNN models like GoogLENet \cite{szegedy2015going} and ResNet \cite{he2016deep}, fine-tuning them on 3D datasets for individual view feature extraction.
However, these CNN models have a large number of parameters, with even the lighter ResNet18 having 11.7 million (M) parameters. Therefore, we compress the CNN structure of GMViT, specifically ResNet18. As illustrated in Table \ref{tab1}, we directly connect multiple 2D convolutional modules and pooling functions without incorporating any residual structures. Following each of these convolutional structures, BatchNorm2d and ReLU activation functions are applied. Moreover, both the view-level ViT and the group-level ViT in GMViT consist of six ViT layers. Additionally, we compress the view-level ViT and group-level ViT as well. GMViT-simple reduces the number of ViT layers to 1 and sets the hidden layer's expansion ratio to 1, whereas GMViT-mini replaces these two models with two minimalist MLPs directly. By compressing the models, GMViT-simple and GMViT-mini, the size of the original large model is reduced from 44.1 M to 5.5 M and 2.5 M, respectively.

\begin{table}[t]
\scriptsize
\centering
\setlength{\tabcolsep}{6pt}
\caption{Network structures of GMViT-simple and GMViT-mini}
\begin{tabular}{|c|c|c|c|}
\hline
\begin{tabular}[c]{@{}c@{}}Network\\Components\end{tabular} & Layer                     & \begin{tabular}[c]{@{}c@{}}Structure\\ Parameter\end{tabular} & \begin{tabular}[c]{@{}c@{}}Activation\\ Function\end{tabular} \\ \hline
\multirow{11}{*}{CNN}             & \multirow{2}{*}{conv2d}   & 7×7, (3, 64),                                                 & \multirow{2}{*}{ReLU}                                         \\
                                  &                           & padding 3, stride 2×2                                         &                                                               \\ \cline{2-4} 
                                  & \multirow{2}{*}{pooling1} & MaxPool2d, 3×3                                                & \multirow{2}{*}{-}                                            \\
                                  &                           & padding 1, stride 2×2                                         &                                                               \\ \cline{2-4} 
                                  & \multirow{2}{*}{conv2d}   & 3×3, (64, 128),                                               & \multirow{2}{*}{ReLU}                                         \\
                                  &                           & stride 2×2                                                    &                                                               \\ \cline{2-4} 
                                  & \multirow{2}{*}{conv2d}   & 3×3, (128, 256)                                               & \multirow{2}{*}{ReLU}                                         \\
                                  &                           & stride 2×2                                                    &                                                               \\ \cline{2-4} 
                                  & \multirow{2}{*}{conv2d}   & 3×3, (256, 512)                                               & \multirow{2}{*}{ReLU}                                         \\
                                  &                           & stride 2×2                                                    &                                                               \\ \cline{2-4} 
                                  & pooling2                  & global  average pooling                                       & -                                                             \\ \hline
encoder1(mini)                    & mlp                       & (512, 512)                                                    & -                                                             \\ \cdashline{1-4}[3pt/2pt]
\multirow{2}{*}{encoder1(simple)} & \multirow{2}{*}{ViT}      & head 8, layer1,                                               & \multirow{2}{*}{-}                                            \\
                                  &                           & mlp hidden dim 512                                          &                                                               \\ \hline
grouping module                     & mlp                       & (512, 1)                                                      & sigmoid                                                       \\ \hline
encoder2(mini)                    & mlp                       & (512, 512)                                                    & -                                                             \\ \cdashline{1-4}[3pt/2pt]
\multirow{2}{*}{encoder2(simple)} & \multirow{2}{*}{ViT}      & head 8, layer1,                                               & \multirow{2}{*}{-}                                            \\
                                  &                           & mlp hidden dim 512                                          &                                                               \\ \hline
pooling                     & max-pooling                       & -                                                      & -                                                      \\ \hline
\multirow{3}{*}{\begin{tabular}[c]{@{}c@{}}classification   \\ head\end{tabular}} & mlp                       & (512, 512)                                                    & ReLU                                                          \\ \cline{2-4}
& mlp                       & (512, 256)                                                    & ReLU                                                          \\ \cline{2-4} 
& mlp                       & (256, num\_class)                                             & ReLU                                                          \\ \hline 
\end{tabular}
\label{tab1}
\end{table}
\subsection{Knowledge distillation}
\begin{figure*}[t]
	\centering
	\includegraphics[width=0.8\linewidth]{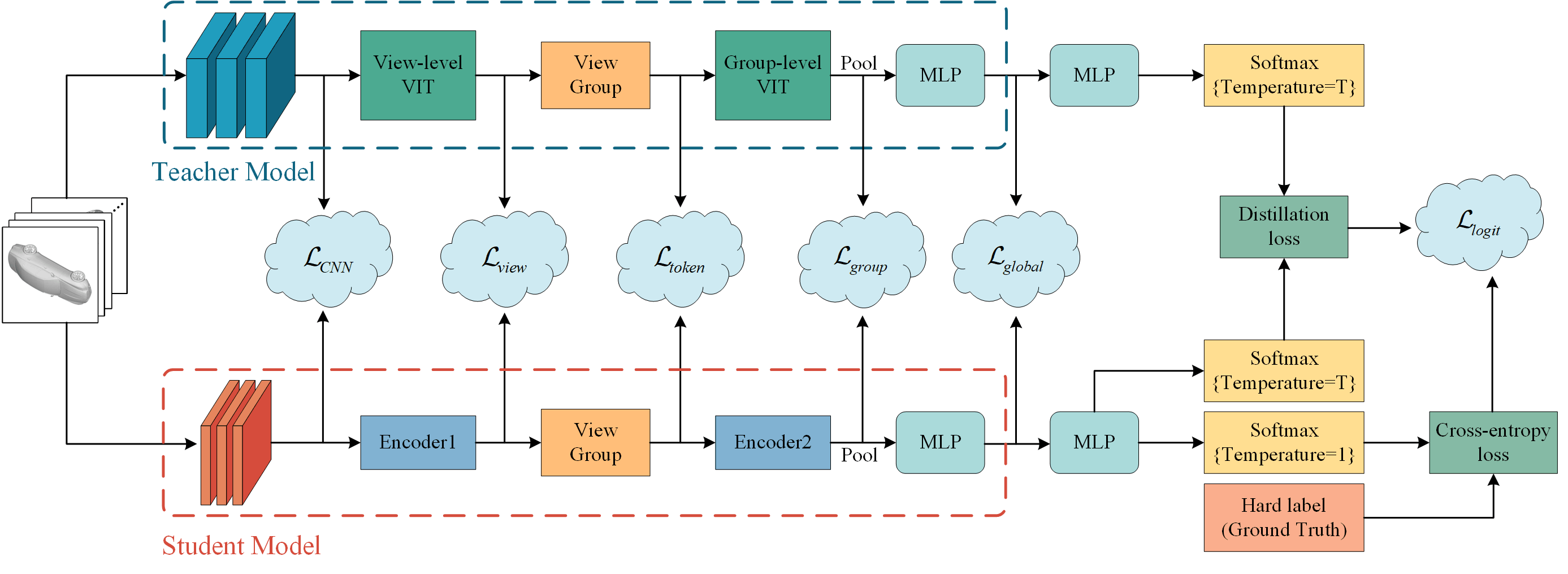}
	\caption{Flow chart of knowledge distillation for GMViT. The output of each component of GMViT is used as the distillation target.}
	\label{fig2}
\end{figure*}

In this section, we employ the knowledge distillation method to enhance the training effectiveness of the small model. This method involves using the output knowledge of the pre-trained large model as the learning target for the small model. During the distillation process, the more powerful GMViT model serves as the teacher model, while the GMViT-simple and GMViT-mini models, which are weaker but refined, act as the student models. As illustrated in Fig. \ref{fig2}, this section employs a comprehensive distillation approach throughout the large model to preserve its performance to the fullest extent. The distillation process includes CNN feature distillation, view-level ViT feature distillation, group token distillation, group-level ViT feature distillation, global feature distillation, and prediction-logit distillation.

\subsubsection{CNN feature distillation} Recent studies have demonstrated that distilling the output of the network's middle layer enhances the training effectiveness, validating the feature-based distillation is reasonable. The CNN module utilized in GMViT primarily consists of ResNet18, which has a well-designed structure, enabling it to effectively learn view features. We employ the mean square error (MSE) between the output $F_t^{CNN}$ of the teacher CNN and the output $F_s^{CNN}$ of the student CNN as the distillation target: 
\begin{equation}
	\mathcal{L}_{CNN}=(1/N)\sum_{n=1}^{N}{MSE(F_{t_n}^{CNN\ },\ F_{s_n}^{CNN})}
	\label{eqa19}
\end{equation}

\subsubsection{View-level ViT feature distillation} The view-level ViT leverages deep ViTs to make the view features fully interactive and strengthen global relationships. The Encoder1 in the student model corresponds to a simple MLP or a single-layer lightweight ViT, which has limited capability in capturing view relations. Hence, we distill the superior view-level features 
$F_{t}^{view}$ learned by the teacher model into the Encoder1 of the student model. The distillation target is defined as follows:
\begin{equation}
	\mathcal{L}_{view}=(1/N)\sum_{n=1}^{N}{MSE(F_{t_n}^{view\ },\ F_{s_n}^{view})}
	\label{eqa20}
\end{equation}
where $F_{s}^{view}$ represents the output feature of the Encoder1.

\subsubsection{Group token distillation} The grouping module of the teacher network has undergone thorough training and demonstrates effective grouping of upper-level features $F_{t}^{view}$. As the previous distillations have significantly aligned $F_{t}^{view}$ and $F_{s}^{view}$, the group token ${Token}_t$ from the teacher network can also be transferred to the student network. Therefore, the distillation target is defined as follows: 
\begin{equation}
    \mathcal{L}_{token}=MSE({Token}_t,\ {Token}_s)
    \label{eqa21}
\end{equation}
where ${Token}_s$ represents the group token of student model.

\subsubsection{Group-level ViT feature distillation} Similarly, we take the MSE of the group-level ViT output feature $F_{t}^{group}$ and the Encoder2 output feature $F_{s}^{group}$ as the optimization target:
\begin{equation}
	\mathcal{L}_{group}=(1/M)\sum_{m=1}^{M}{MSE(F_{t_m}^{group},\ F_{s_m}^{group})}
	\label{eqa22}
\end{equation}

\subsubsection{Global feature distillation} We use the shape descriptor $F_D$ in Equation \ref{eqa15} as the global feature. As the global features are utilized directly in downstream tasks, distilling the global features becomes essential: 
\begin{equation}
	\mathcal{L}_{global}=MSE(F_t^{global},\ F_s^{global})
	\label{eqa23}
\end{equation}
where $F_t^{global}$ and $F_s^{global}$ represent the shape descriptors of the teacher model and the student model, respectively.

\subsubsection{Prediction-logit distillation} Hinton et al. \cite{hinton2015distilling} employed the soft label ${pred}_t$, derived from the output of the teacher model, as the distillation target for optimizing the student model. They demonstrated that this approach is more effective in enhancing model performance compared to traditional training methods. Consequently, we incorporate the soft label loss $\mathcal{L}_{soft}$ into the prediction loss. Furthermore, we introduce the hard label loss $L_{hard}$, which represents the cross-entropy loss between the predicted ${pred}_s$ from the student model and the true labels. Since even the powerful teacher model cannot guarantee the correctness of all predictions, the true labels play a role in correcting errors when needed: 
\begin{equation}
	\mathcal{L}_{soft}=KL(softmax(\frac{{pred}_t}{T}),\ softmax(\frac{{pred}_s}{T}))
	\label{eqa24}
\end{equation}
\begin{equation}
	\mathcal{L}_{hard}=CE(softmax(label),\ softmax({pred}_s))
	\label{eqa25}
\end{equation}
\begin{equation}
	\mathcal{L}_{logit}=(1-\lambda)L_{soft}+\lambda L_{hard}
	\label{eqa26}
\end{equation}
where $KL$ denotes Kullback-Leibler divergence loss, $log\_softmax$ denotes logarithm after passing the softmax function, $T$ denotes distillation temperature, and $CE$ denotes cross-entropy loss.

To sum up, the final distillation target is:
\begin{equation}            
	\mathcal{L}=\mathcal{L}_{CNN}+\mathcal{L}_{view}+\mathcal{L}_{token}+\mathcal{L}_{group}+\mathcal{L}_{global}+\mathcal{L}_{logit}
	\label{eqa27}
\end{equation}

\section{Experiment}
\label{sec4}
\subsection{Datasets}
\textbf{ModelNet:} ModelNet \cite{wu20153d} contains 127,000+ 3D CAD models from 662 categories. ModelNet40 includes 12311 objects from 40 categories (9843/2468 in training/testing). ModelNet10 has 4899 objects from 10 classes (3991/908 for training/testing). We use Circle-12 and Dodecahedron-20 camera settings \cite{xu2021multi} for evaluation. 

\textbf{ShapeNetCore55:} ShapeNetCore55 \cite{chang2015shapenet} is a subset of ShapeNet, containing 51,300 3D objects from 55 categories and 203 subcategories. It's split into a 7:1:2 ratio for training, validation, and testing. We evaluate on the NORMAL version, where 3D objects are aligned. 

\textbf{MCB:} MCB \cite{kim2020large} is a 3D machine part dataset with two versions. MCB-A has 58,696 objects from 68 categories, while MCB-B has 18,038 objects from 25 categories of MCB-A. Objects are sourced from TraceParts, 3D Warehouse, and GrabCAD, without alignment. 

\begin{table*}[t]
	\centering
	\caption{Performance comparison on ModelNet dataset. Bold represents the best results. The $n$× after the method name represents the number of input views. `KD' represents training the model by knowledge distillation.}
	\begin{tabular}{|c|c|cccccc|}
		\hline
		\multirow{3}{*}{Input}                 & \multirow{3}{*}{Method} & \multicolumn{3}{c|}{ModelNet40}                                                            & \multicolumn{3}{c|}{ModelNet10}                                             \\ \cline{3-8} 
		&                         & \multicolumn{2}{c|}{Classification}                       & \multicolumn{1}{c|}{Retrieval} & \multicolumn{2}{c|}{Classification}                       & Retrieval      \\ \cline{3-8} 
		&                         & \multicolumn{1}{c|}{OA(\%)} & \multicolumn{1}{c|}{mA(\%)} & \multicolumn{1}{c|}{mAP(\%)}   & \multicolumn{1}{c|}{OA(\%)} & \multicolumn{1}{c|}{mA(\%)} & mAP(\%)        \\ \hline
		\multirow{2}{*}{Voxels}                & 3D ShapeNet \cite{wu20153d}             & 77.32                       & -                           & 49.32                          & 83.54                       & -                           & 68.26          \\
		& VRN-Ensemble \cite{brock2016generative}            & 95.54                       & -                           & -                              & 97.14                       & -                           & -              \\ \hline
		\multirow{5}{*}{Point Cloud}                                                  & PointNet \cite{qi2017pointnet}                                    & 89.20           & 86.20                        & -                              & -              & -                           & -              \\
		& PointNet++ \cite{qi2017pointnet++}                                  & 91.90           & -                           & -                              & -              & -                           & -              \\
		& point2vec \cite{zeid2023point2vec}                                       & 94.80           & 92.00                        & -                           & -              & -                           & -              \\
		& PointMLP \cite{ma2022rethinking}                                      & 94.50           & 91.40                        & -                              & -           & -                        & -              \\ 
		& GeomGCNN \cite{srivastava2021exploiting}                                       & 95.90         & 93.10                        & -                           & -              & -                           & -              \\    \hline
		\multirow{2}{*}{Point Cloud and Views} & PVNet \cite{you2018pvnet}, 12×              & 93.2                        & 91                          & 89.5                           & -                           & -                           & -              \\
		& PVRNet \cite{you2019pvrnet},   12×           & 93.61                       & 91.64                       & 90.5                           & -                           & -                           & -              \\ \hline
		\multirow{24}{*}{Views}                & MVCNN \cite{su2015multi}, 80×              & 90.1                        & -                           & 79.5                           & -                           & -                           & -              \\
		& GVCNN \cite{feng2018gvcnn}, 8×               & 93.1                        & 90.7                        & 85.7                           & -                           & -                           & -              \\
		& GIFT \cite{bai2016gift}, 64×               & -                           & 89.5                        & 91.94                          & -                           & 91.5                        & 91.12          \\
		& MHBN \cite{yu2018multi}, 6×                & 94.1                        & 92.2                        & -                              & 94.9                        & 94.9                        & -              \\
		& 3D2SeqViews \cite{han20193d2seqviews}, 12×        & 93.4                        & 91.51                       & 90.76                          & 94.71                       & 94.68                       & 92.12          \\
		& Ma et al. \cite{ma2018learning}, 12×          & 91.05                       & -                           & 84.34                          & 95.29                       & -                           & 93.19          \\
		& DAN \cite{nie2021dan}, 12×                & 93.5                        & -                           & 90.4                           & 94.9                        & -                           & 92.3           \\
		& RelationNet \cite{yang2019learning}, 12×        & 94.3                        & 92.3                        & 86.7                           & 95.3                        & 95.1                        & -              \\
		& RotationNet \cite{kanezaki2018rotationnet}, 20×        & 97.37                       & 94.68                       & -                              & 98.46                       & 94.82                       & -              \\
		& View-GCN \cite{wei2020view}, 20×           & 97.6                        & 96.5                        & -                              & -                           & -                           & -              \\
		& CAR-Net \cite{xu2021multi}, 12×            & 95.22                       & -                           & 91.27                          & 95.82                       & -                           & 91.53          \\
		& CAR-Net, 20×            & 97.73                       & -                           & 95.04                          & \textbf{99.01}                       & -                           & 97.12          \\ \cline{2-8} 
		& GMViT(Ours), 12×        & 96.27                       & 93.99                       & 94.54                          & 98.79                       & 98.7                        & 98.35          \\
		& GMViT(Ours), 20×        & \textbf{97.77}              & \textbf{97.07}              & \textbf{97.57}                 & \textbf{99.01}              & \textbf{98.92}              & \textbf{98.63} \\
		& GMViT-simple, 12×       & 91.9                        & 88.86                       & 86.19                          & 92.62                       & 92.3                        & 86.79          \\
		& GMViT-simple(KD), 12×   & 92.95\scriptsize{\textbf{(+1.05)}}                & 89.62\scriptsize{(\textbf{+0.76})}                & 90.54\scriptsize{(\textbf{+4.35})}                   & 97.03\scriptsize{(\textbf{+4.4})}                 & 96.97\scriptsize{(\textbf{+4.67})}                & 95.72\scriptsize{(\textbf{+8.93})}   \\
		& GMViT-mini, 12×         & 89.55                       & 86.01                       & 80.88                          & 91.96                       & 91.97                       & 86.59          \\
		& GMViT-mini(KD), 12×     & 92.42\scriptsize{(\textbf{+2.87})}                & 88.99\scriptsize{(\textbf{+2.98})}                & 85.84\scriptsize{(\textbf{+4.96})}                   & 94.71\scriptsize{(\textbf{+2.75})}                & 94.44\scriptsize{(\textbf{+2.47})}                & 91.82\scriptsize{(\textbf{+5.23})}   \\
		& GMViT-simple, 20×       & 95.06                       & 92.82                       & 89.44                          & 98.35                       & 98.2                        & 97.38          \\
		& GMViT-simple(KD), 20×   & 95.75\scriptsize{(\textbf{+0.69})}                & 93.55\scriptsize{(\textbf{+0.73})}                & 94.24\scriptsize{(\textbf{+4.8})}                    & 98.46\scriptsize{(\textbf{+0.11})}                & 98.42\scriptsize{(\textbf{+0.22})}                & 98.14\scriptsize{(\textbf{+0.76})}   \\
		& GMViT-mini, 20×         & 93.44                       & 89.91                       & 87.36                          & 97.91                       & 97.94                       & 95.94          \\
		& GMViT-mini(KD), 20×     & 95.75\scriptsize{(\textbf{+2.31})}                & 92.41\scriptsize{(\textbf{+2.5})}                 & 91.12\scriptsize{(\textbf{+3.76})}                   & 98.79\scriptsize{(\textbf{+0.88})}                & 98.62\scriptsize{(\textbf{+0.68})}                & 97.14\scriptsize{(\textbf{+1.2})}    \\ \hline
	\end{tabular}
	\label{tab2}
\end{table*}
\subsection{Implementation details}

Each 3D object is rendered into 224 × 224 2D images. GMViT's CNN backbone is based on ResNet18 \cite{he2016deep}, excluding the last fully connected layer. Both the view-level ViT and group-level ViT have 6 layers and 8 attention heads each. The grouping module is set with 8 and 12 groups for Circle-12 and Dodecahedron-20 settings, respectively. GMViT-simple and GMViT-mini use the same grouping module settings as the large model. A Dropout layer with a 0.5 dropout rate is added to address overfitting.

The model is trained using the SGD optimizer with 1e-4 momentum and weight decay for 100 epochs. The learning rate starts at 0.1 and decreases to 0.01 over 50 epochs with cosine annealing. Different strategies are used for training large and small models. Large model CNNs are pre-trained on ImageNet before fine-tuning on the 3D shape dataset. Small model CNNs are directly integrated during training. The distillation temperature is set to 5.

\subsection{Experiments on ModelNet}
In this section, we present the classification and retrieval performance analysis of the proposed model on the ModelNet dataset. To validate the effectiveness of our proposed method, we compare it with a wide range of methods, including voxel-based (3D ShapeNet \cite{wu20153d} and VRN-Ensemble \cite{brock2016generative}), point-based (PointNet \cite{qi2017pointnet}, PointNet++ \cite{qi2017pointnet++}, GeomGCNN \cite{srivastava2021exploiting}, point2vec \cite{zeid2023point2vec} and PointMLP \cite{ma2022rethinking}), multimodal-based (PVNet \cite{you2018pvnet} and PVRNet \cite{you2019pvrnet}), and view-based (MVCNN \cite{su2015multi}, GVCNN \cite{feng2018gvcnn}, GIFT \cite{bai2016gift}, MHBN \cite{yu2018multi}, 3D2SeqViews \cite{han20193d2seqviews}, Ma et al. \cite{ma2018learning}, DAN \cite{nie2021dan}, 
RelationNet \cite{yang2019learning}, CAR-Net \cite{xu2021multi}, RotationNet \cite{kanezaki2018rotationnet}, and View-GCN \cite{wei2020view}) approaches. The primary evaluation metrics for classification are overall accuracy (OA) and mean class accuracy (mA). 
For the retrieval task, the shape descriptors are obtained by directly utilizing the 256-dimensional features from the classifier's penultimate fully connected layer. In the retrieval task, each object in the testing set is treated as a query, and a KD-Tree is employed to rank the similarity of its feature to the remaining object features. The mean average precision (mAP) is subsequently calculated based on this ranking.

\begin{table*}[t]
\centering
\caption{Performance comparison on ShapenetCore55 dataset. Bold represents the best results.}
\setlength{\tabcolsep}{11pt}
\begin{tabular}{|l|ccccc|ccccc|}
\hline
\multirow{2}{*}{Methods} & \multicolumn{5}{c|}{microALL}                                                 & \multicolumn{5}{c|}{macroALL}                                               \\
                         & P@N           & R@N           & F1@N          & mAP           & NDCG          & P@N           & R@N           & F1@N          & mAP           & NDCG        \\ \hline
ZFDR                     & 53.5          & 25.6          & 28.2          & 19.9          & 33.0            & 21.9          & 40.9          & 19.7          & 25.5          & 37.7        \\
DeepVoxNet               & 79.3          & 21.1          & 25.3          & 19.2          & 27.7          & 59.8          & 28.3          & 25.8          & 23.2          & 33.7        \\
DLAN                     & 81.8          & 68.9          & 71.2          & 66.3          & 76.2          & \textbf{61.8} & 53.3          & 50.5          & 47.7          & 56.3        \\
GIFT                     & 70.6          & 69.5          & 68.9          & 64.0            & 76.5          & 44.4          & 53.1          & 45.4          & 44.7          & 54.8        \\
Improved GIFT            & 78.6          & 77.3          & 76.7          & 72.2          & 82.7          & 59.2          & \textbf{65.4} & 58.1          & 57.5          & 65.7        \\
ReVGG                    & 76.5          & 80.3          & 77.2          & 74.9          & 82.8          & 51.8          & 60.1          & 51.9          & 49.6          & 55.9        \\
MVFusionNet              & 74.3          & 67.7          & 69.2          & 62.2          & 73.2          & 52.3          & 49.4          & 48.4          & 41.8          & 50.2        \\
CM-VGG5-6DB              & 41.8          & 71.7          & 47.9          & 54.0            & 65.4          & 12.2          & 66.7          & 16.6          & 33.9          & 40.4        \\
MVCNN                    & 77.0            & 77.0            & 76.4          & 73.5          & 81.5          & 57.1          & 62.5          & 57.5          & 56.6          & 64.0          \\
RotationNet              & 81.0            & 80.1 & 79.8          & 77.2          & 86.5 & 60.2          & 63.9          & 59.0            & 58.3          & 65.6        \\
MVCNN(VAM+IAM)           & -            & - & 79.9          & \textbf{80.9}          & 86.7 & -          & -          & 59.3            & \textbf{63.0}          & \textbf{66.7}        \\
GMViT(Ours)              & \textbf{81.3} & \textbf{80.9} & \textbf{80.7} & 77.5 & \textbf{86.9}          & 61.3          & 65.1          & \textbf{60.2} & 60.5 & \textbf{66.7} \\ \hline
\end{tabular}
\label{tab3}
\end{table*}
\subsubsection{Classification results}
Table \ref{tab2} shows the model's classification performance. Generally, view-based methods outperform point-based methods significantly. Our GMViT achieves optimal performance across all indicators in both datasets under the Dodecahedron-20 setting. Our GMViT demonstrates an approximate 2\% improvement in OA for both datasets compared to the optimal voxel-based model VRN-Ensemble \cite{brock2016generative}. Our GMViT achieves comparable classification performance to the current leading view-based method, CAR-Net  \cite{xu2021multi}. Both methods consider the spatial relationship of views from different perspectives. Our method also outperforms other methods in the Circle-12 setting alone. DAN \cite{nie2021dan} replaces parallel multi-head self-attention with deep self-attention to enhance the fusion of significant 3D features between views. In contrast, our approach incorporates the spatial information of views in the position embedding part of GMViT and extends the consideration beyond the relationship between views to encompass the relationship between groups. 3D2SeqViews \cite{han20193d2seqviews} considers a view as a sequence and captures its dependencies through hierarchical attention aggregation. However, this approach largely overlooks the positional relationships of views in 3D space. Unlike 3D2SeqViews, we utilize the rendering coordinates of the view as the position embedding , enabling us to convert the 1D sequence relations into their corresponding 3D spatial relations. GVCNN \cite{feng2018gvcnn} incorporates group pooling before global pooling, resulting in reduced pooling scale and effectively mitigating feature loss. In comparison to GVCNN, we introduce two types of ViT for establishing the relationship between view-level and group-level features before group pooling and global pooling, respectively. This approach further minimizes feature loss attributed to pooling.

\begin{figure}[t]
	\centering
	\subfloat[GMViT]{\includegraphics[width = 0.14925\textwidth]{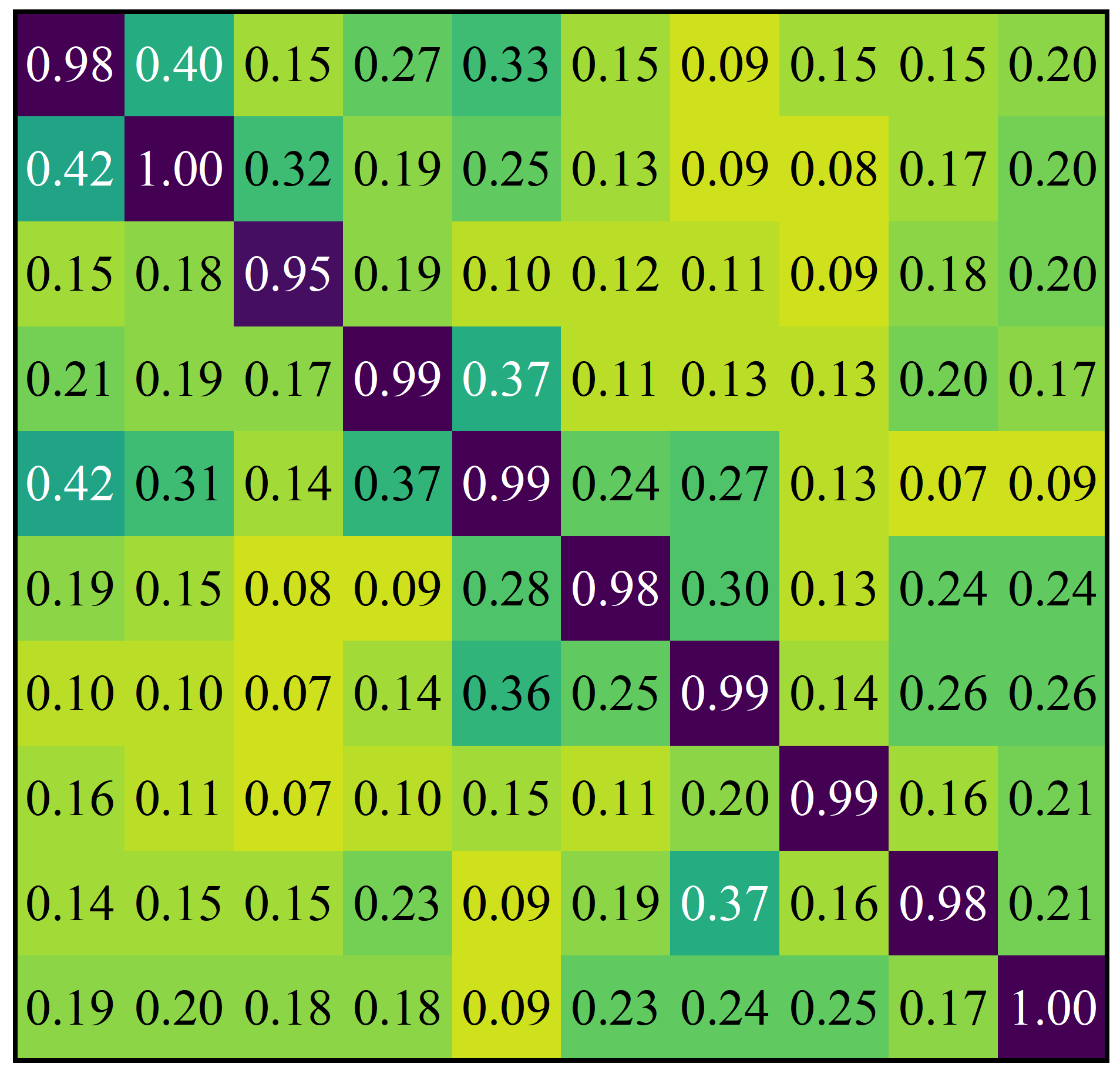}}
	\hfill
	\subfloat[GMViT-mini]{\includegraphics[width = 0.15\textwidth]{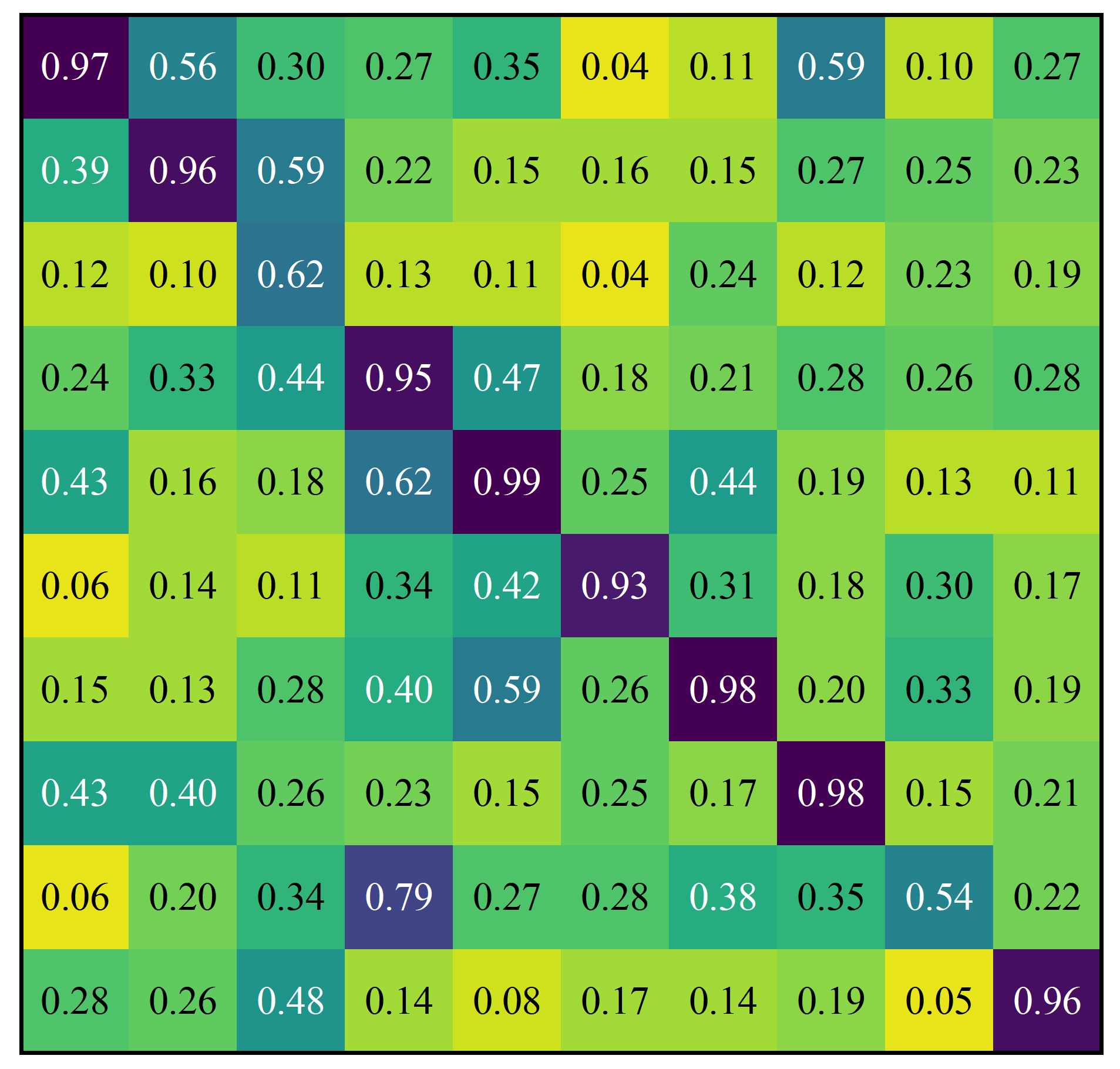}}
	\hfill
	\subfloat[GMViT-mini(KD)]{\includegraphics[width = 0.15\textwidth]{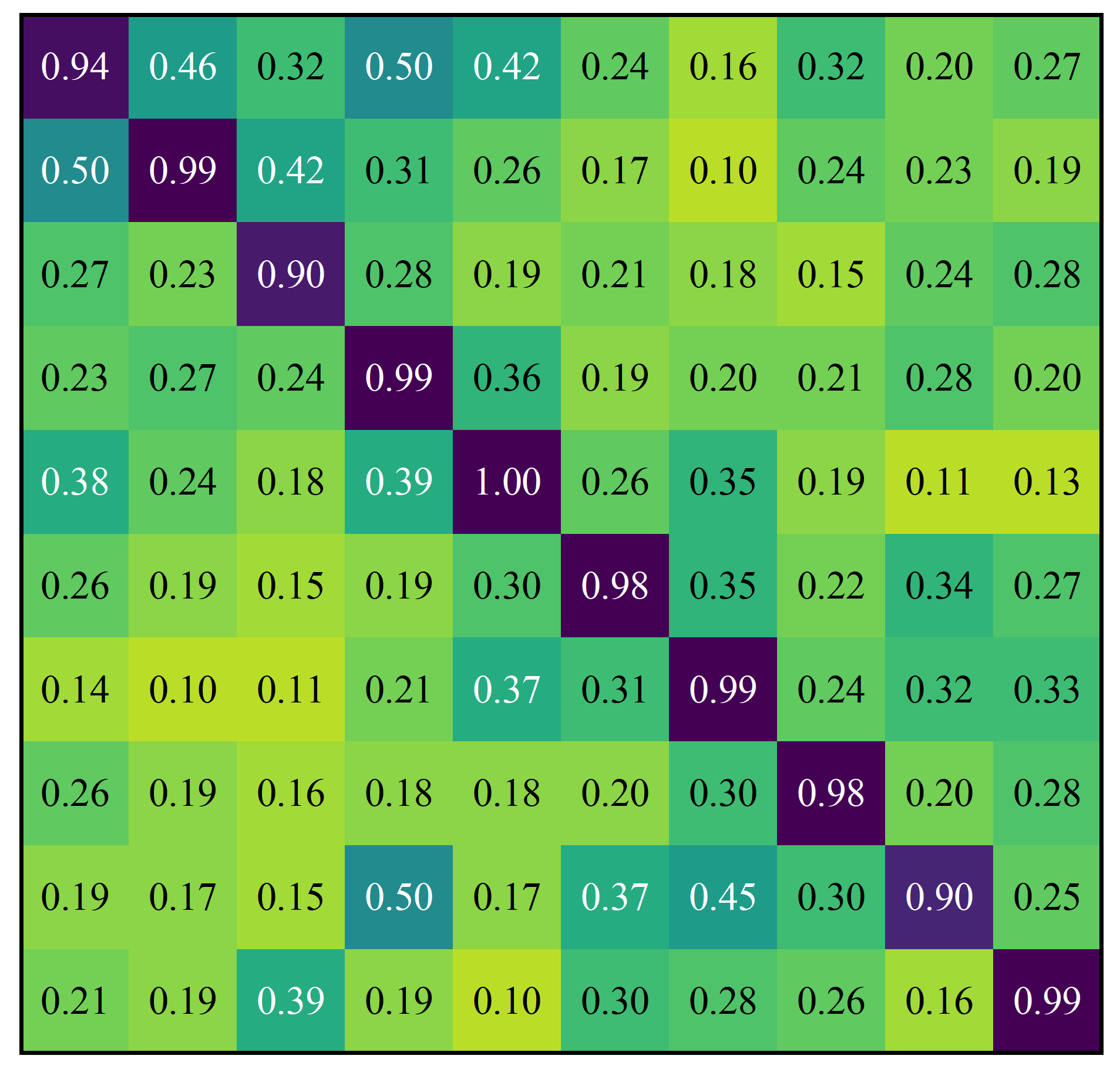}}
	\hfill
	\subfloat[GMViT-simple]{\includegraphics[width = 0.15\textwidth]{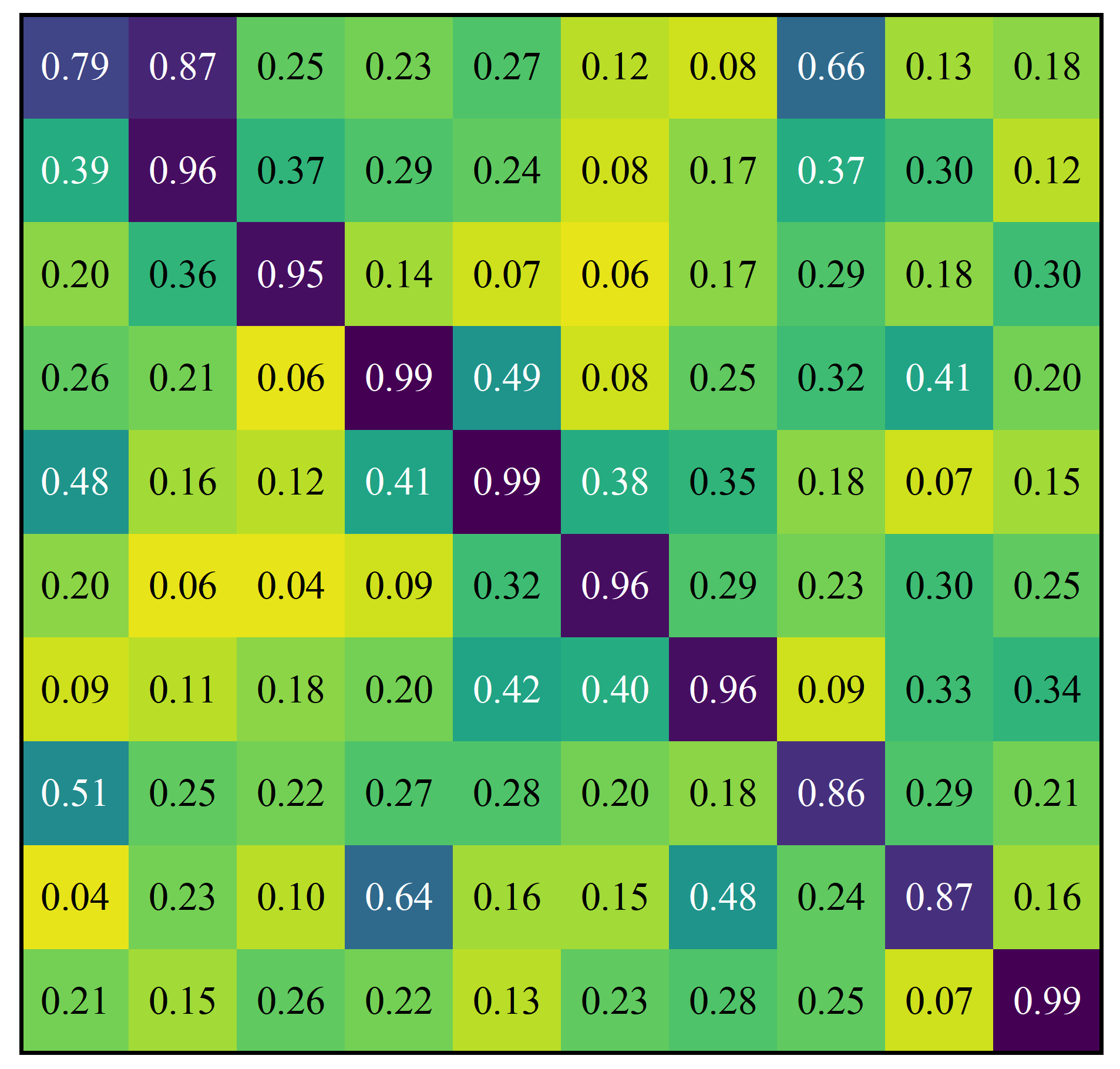}}
	\hspace {10mm}
	\subfloat[GMViT-simple(KD)]{\includegraphics[width = 0.15\textwidth]{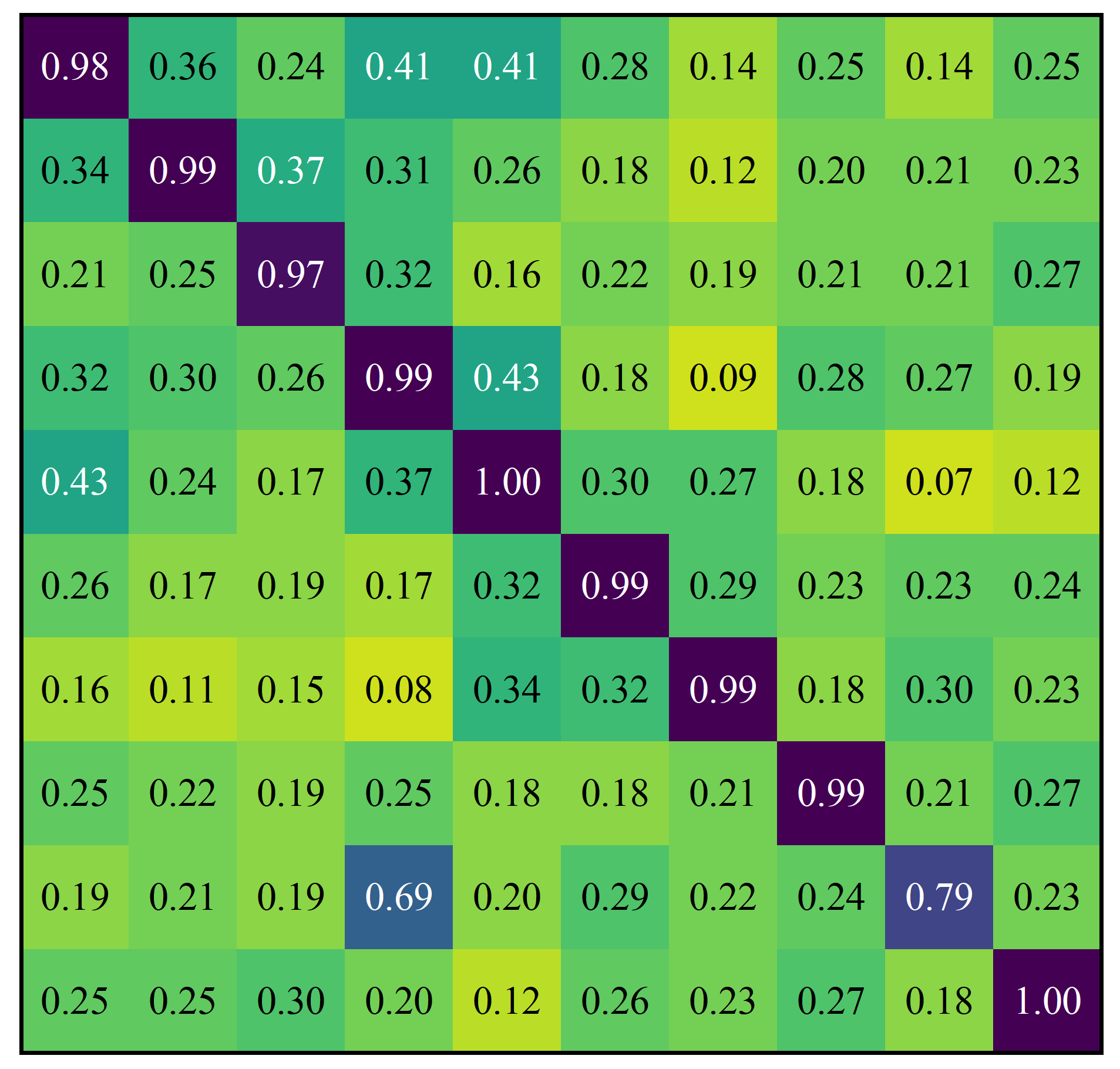}}
	\caption{Similarity of the 3D shape descriptors learned by the five proposed models. The two objects with the same row and column numbers on each heat map are from the same category of ModelNet10. Each heat map includes ten categories of objects.}
	\label{fig3}
\end{figure}
In addition, we evaluate the performance of the small models GMViT-simple and GMViT-mini. Directly training small models with hard labels leads to unsatisfactory classification results due to performance degradation resulting from simplified networks. However, the small models trained using our proposed knowledge distillation method are more effectively optimized. Significantly, the student model achieves classification performance on the ModelNet10 dataset that is comparable to the current state-of-the-art method, CAR-Net.

\subsubsection{Retrieval results}
Regarding shape retrieval, our GMViT also demonstrates outstanding performance. While GMViT and CARNet achieved comparable classification results under the Dodecahedron-20 setting, GMViT outperforms CARNet in retrieval performance with improvements of 2.53\% and 1.51\% on the respective datasets. Conversely, CAR-Net's \cite{xu2021multi} retrieval performance is not superior to that of all other methods under the Circle-12 setting. Remarkably, our GMViT outperforms all other methods, including Dodecahedron-20, on ModelNet10 while maintaining superiority under the same setting. This provides evidence of the superior ability of our proposed GMViT to learn more effective 3D shape descriptors.

Similarly, GMViT-simple and GMViT-mini demonstrate substantial and comprehensive improvements in retrieval performance following the distillation process. Particularly noteworthy, GMViT-simple and GMViT-mini outperform all other large models on the ModelNet10 dataset when evaluated under the Dodecahedron-20 setting. This remarkable achievement can primarily be attributed to the exceptional retrieval performance of the student models, which is inherited from the teacher model through the distillation process. To better show the similarity of the 3D shape descriptors learned by each model, we plot them in the Fig. \ref{fig3}. 

\subsection{Experiments on ShapeNetCore55}
In order to comprehensively evaluate the shape retrieval performance of GMViT, we conduct experiments on the ShapeNetCore55 dataset. Consistent with the experimental setup described in \cite{savva2017large}, we limit the retrieval to a maximum of 1000 shapes per query. In the retrieval process, we utilize multiple views as model input under the Dodecahedron-20 setting and employed KD-Tree to generate the retrieval score ranking for each shape. We utilize indicators under both ``microALL" and ``macroALL" settings. ``microALL" represents a weighted average based on the category size of the samples, while ``macroALL" does not consider such weighting. The retrieval results, sourced from \cite{savva2017large}, are presented in Table \ref{tab3}. Our method's performance is only slightly lower than the competition-winning method, RotationNet, in terms of the NDCG indicator under the ``microALL" setting. In the retrieval task, P@N and R@N indicators demonstrate a trade-off relationship. Our method achieves a better balance between P@N and R@N compared to the runner-up method, DLAN. Compared with the recently published method MVCNN(VAM+IAM) \cite{lin2022multi}, our GMViT demonstrates greater advantages in general.

\begin{table*}[t]
\centering
\caption{Retrieval comparison on the MCB-A dataset. The best results are shown in bold.}
\setlength{\tabcolsep}{7.5pt}
\begin{tabular}{|l|ccc|ccc|ccc|}
\hline
\multirow{2}{*}{Methods} & \multicolumn{3}{c|}{microALL}                 & \multicolumn{3}{c|}{macroALL}                  & \multicolumn{3}{c|}{microALL + macroALL}      \\
                         & F1@N          & MAP           & NDCG@N        & F1@N          & MAP            & NDCG@N        & F1@N        & MAP             & NDCG@N        \\ \hline
PointCNN \cite{li2018pointcnn}                 & 69.0            & 88.9          & 89.8          & \textbf{83.3} & 88.6           & 85.4          & 76.2        & 88.3            & 87.6          \\
PointNet++ \cite{qi2017pointnet++}              & 61.3          & 79.4          & 75.4          & 71.2          & 80.3           & 74.6          & 66.3        & 79.9            & 75.0            \\
SpiderCNN \cite{xu2018spidercnn}                & 66.9          & 86.7          & 79.3          & 77.6          & 87.7           & 81.2          & 72.3        & 87.2            & 80.3          \\
MVCNN \cite{su2015multi}                    & 48.8          & 65.7          & 48.7          & 58.5          & 73.5           & 64.1          & 53.7        & 69.6            & 56.4          \\
RotationNet \cite{kanezaki2018rotationnet}              & 50.8          & 80.5          & 68.3          & 68.3          & 81.5           & 73.5          & 56.0          & 81.0              & 70.9          \\
DLAN \cite{furuya2016deep}                     & 56.8          & 87.9          & 82.8          & 82.0            & 88.0             & 84.5          & 69.4        & 88.0              & 83.7          \\
VRN \cite{brock2016generative}                      & 40.2          & 65.3          & 51.9          & 50.7          & 66.4           & 57.6          & 45.5        & 65.9            & 54.8          \\ \hline
GMViT                    & \textbf{92.8} & \textbf{96.5} & \textbf{95.7} & 61.1          & 89.0 & \textbf{87.9} & \textbf{77.0} & 92.7 & \textbf{91.8} \\
GMViT-mini               & 91.6          & 94.7          & 93.8          & 59.3          & 85.1           & 84.7          & 75.5        & 89.9            & 89.3          \\
GMViT-mini(KD)           & 92.6\scriptsize{(\textbf{+1.0})}      & {95.8\scriptsize(\textbf{+1.1})}    & 95.2\scriptsize{(\textbf{+1.4})}    & 60.6\scriptsize{(\textbf{+1.3})}    & 87.3\scriptsize{(\textbf{+2.2})}      & 86.9\scriptsize{(\textbf{+2.2})}    & 76.6\scriptsize{(\textbf{+1.1})}  & 91.6\scriptsize{(\textbf{+1.7})}      & 91.1\scriptsize{(\textbf{+1.8})}    \\
GMViT-simple             & 91.7          & 95.1          & 93.9          & 59.1          & 85.5           & 84.1          & 75.4        & 90.3            & 89            \\
GMViT-simple(KD)         & 92.7\scriptsize{(\textbf{+1.0})}      & 96.4\scriptsize{(\textbf{+1.3})}    & 95.5\scriptsize{(\textbf{+1.6})}    & 61.0\scriptsize{(\textbf{+1.9})}      & \textbf{89.3}\scriptsize{(\textbf{+3.8})}     & 87.9\scriptsize{(\textbf{+3.8})}    & 76.9\scriptsize{(\textbf{+1.5})}  & \textbf{92.9}\scriptsize{(\textbf{+2.6})}      & 91.7\scriptsize{(\textbf{+2.7})}    \\
\hline
\end{tabular}
\label{tab5}
\end{table*}
\begin{table}[t]
\centering
\caption{Classification comparison on the MCB-A dataset. The best results are shown in bold.}
\setlength{\tabcolsep}{10pt}
\begin{tabular}{|l|cc|}
\hline
Method           & OA(\%)         & mA(\%)         \\ \hline
PointCNN \cite{li2018pointcnn}         & 93.89          & 81.85          \\
PointNet++ \cite{qi2017pointnet++}      & 87.45          & 73.68          \\
SpiderCNN \cite{xu2018spidercnn}        & 93.59          & 79.70           \\
MVCNN \cite{su2015multi}           & 64.67          & 80.47          \\
RotationNet \cite{kanezaki2018rotationnet}      & \textbf{97.35} & \textbf{90.79} \\
DLAN \cite{furuya2016deep}             & 93.53          & 82.97          \\
VRN \cite{brock2016generative}              & 93.17          & 80.34          \\ \hline
GMViT            & 96.31          & 90.15          \\
GMViT-mini       & 93.15          & 88.30           \\
GMViT-mini(KD)   & 94.72\scriptsize{(\textbf{+1.57})}   & 89.11\scriptsize{(\textbf{+0.81})}   \\
GMViT-simple     & 93.37          & 88.73          \\
GMViT-simple(KD) & 95.01\scriptsize{(\textbf{+1.64})}   & 89.33\scriptsize{(\textbf{+0.6})}    \\
\hline
\end{tabular}
\label{tab4}
\end{table}
\subsection{Experiments on MCB}
In this section, we conduct additional experiments on the MCB-A dataset to further validate the effectiveness of the proposed method. The experiments encompass both 3D shape classification and retrieval tasks. The primary comparison methods include PointCNN \cite{li2018pointcnn}, PointNet \cite{qi2017pointnet}, SpiderCNN \cite{xu2018spidercnn}, MVCNN \cite{su2015multi}, RotationNet \cite{kanezaki2018rotationnet}, DLAN \cite{furuya2016deep}, and VRN \cite{brock2016generative}. The experimental results for all the aforementioned methods are obtained from \cite{kim2020large}.

\subsubsection{Classification results}
The classification results of models on MCB-A are presented in Table \ref{tab4}. Among the models, RotationNet \cite{kanezaki2018rotationnet}, an advanced multi-view approach, achieves the highest classification results, with our GMViT ranking second. Despite being a view-based method, MVCNN \cite{su2015multi} exhibits the lowest performance. In contrast, our smaller models, GMViT-simple and GMViT-mini, outperform MVCNN significantly and demonstrate further improvement through distillation. This finding validates that, in view-based methods, the quality of the multi-view feature fusion module holds greater significance than that of a single-view feature extraction module.

\begin{figure}[t]
\centering
	\subfloat[GMViT]{\includegraphics[width = 0.15\textwidth]{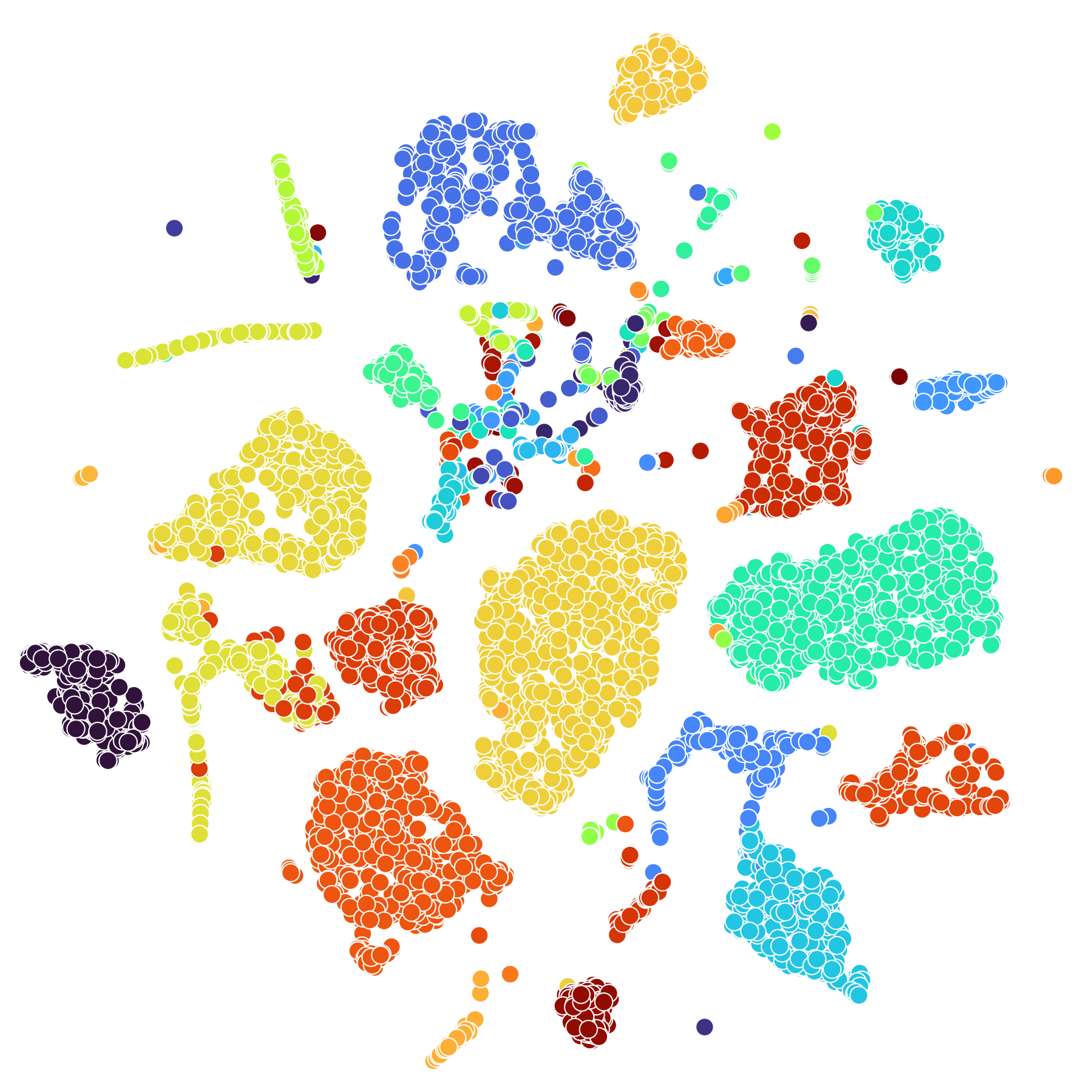}}
	\hfill
	\subfloat[GMViT-mini]{\includegraphics[width = 0.15\textwidth]{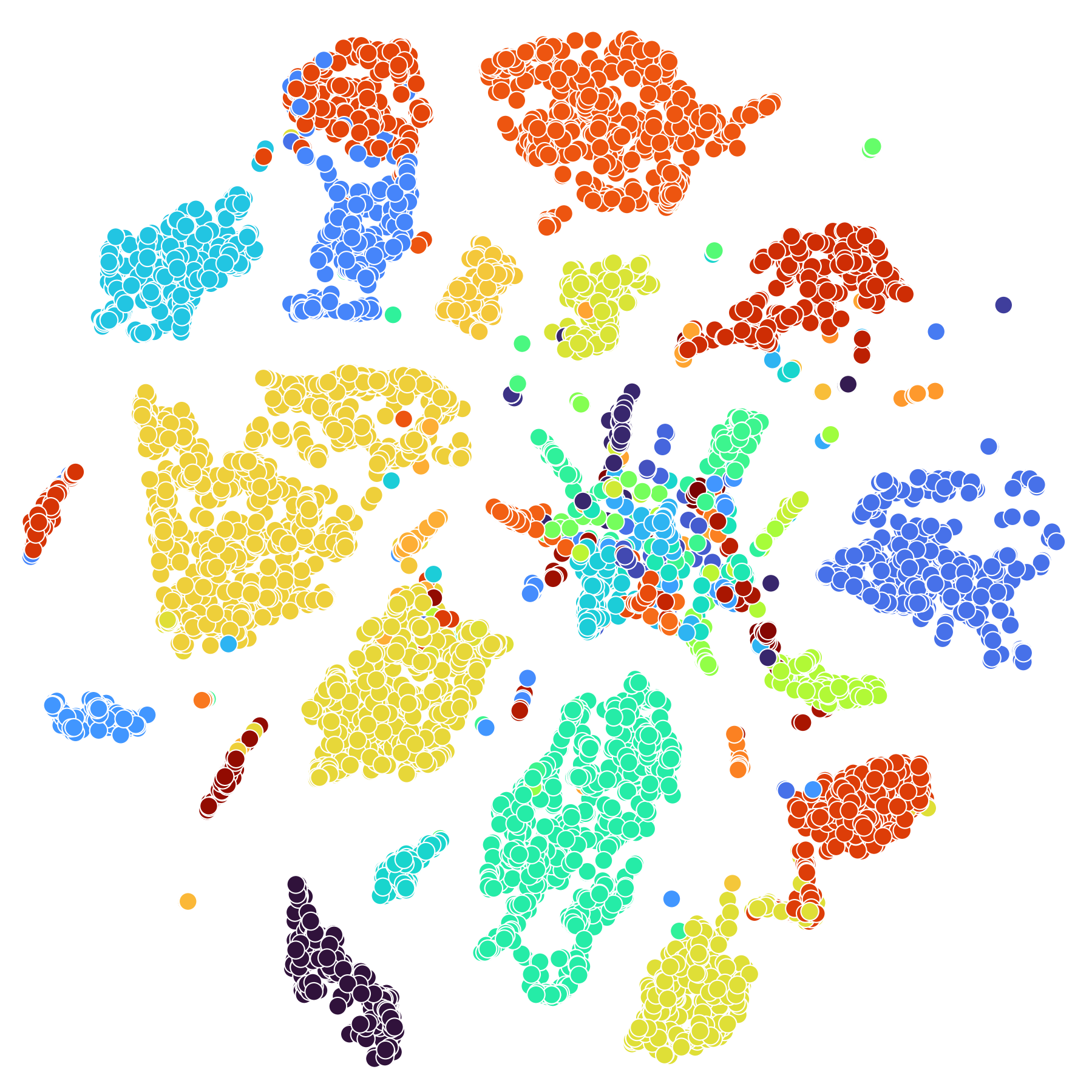}}
        \hfill
	\subfloat[GMViT-mini(KD)]{\includegraphics[width = 0.15\textwidth]{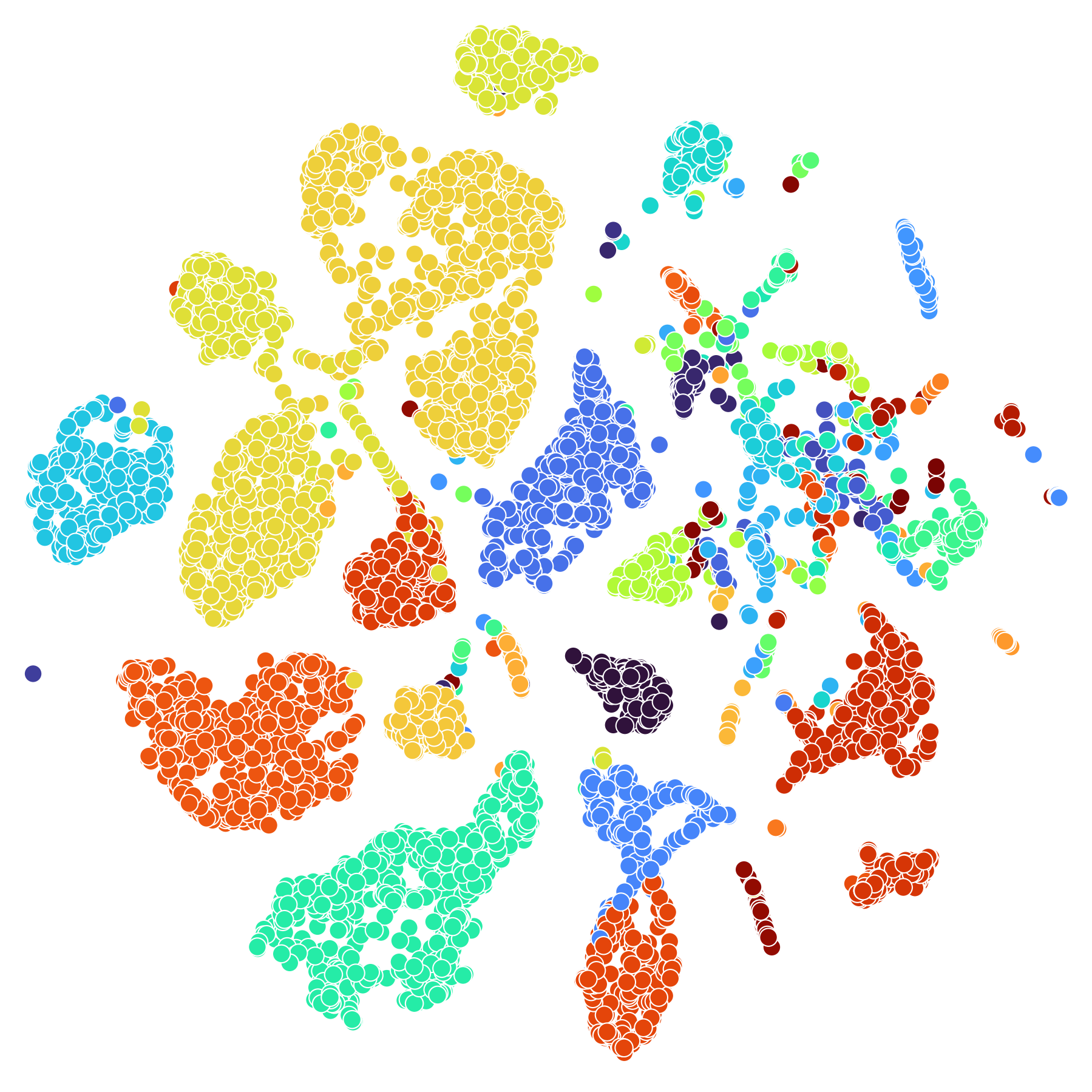}}
        \hfill
	\subfloat[GMViT-simple]{\includegraphics[width = 0.15\textwidth]{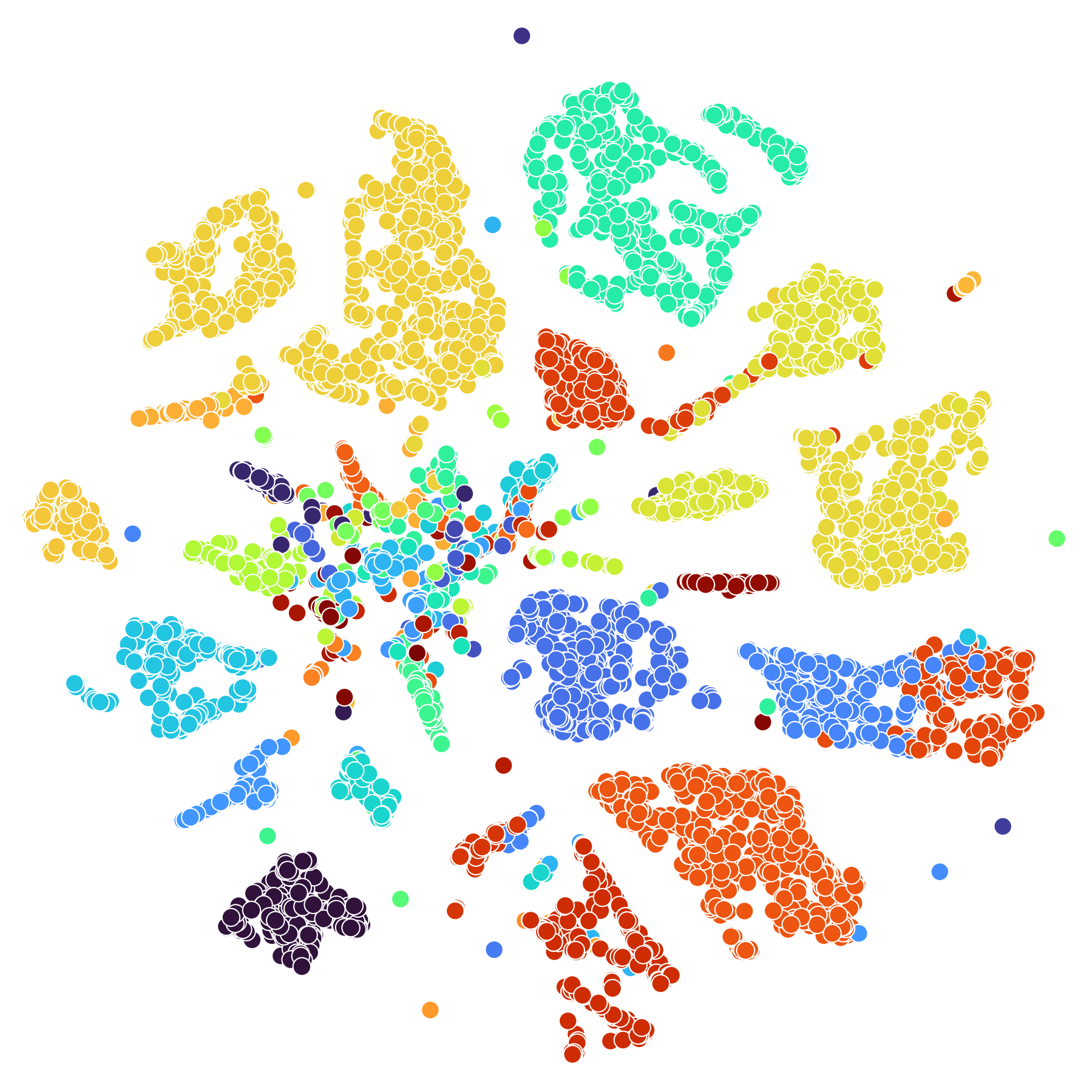}}
        \hspace {10mm}
	\subfloat[GMViT-simple(KD)]{\includegraphics[width = 0.15\textwidth]{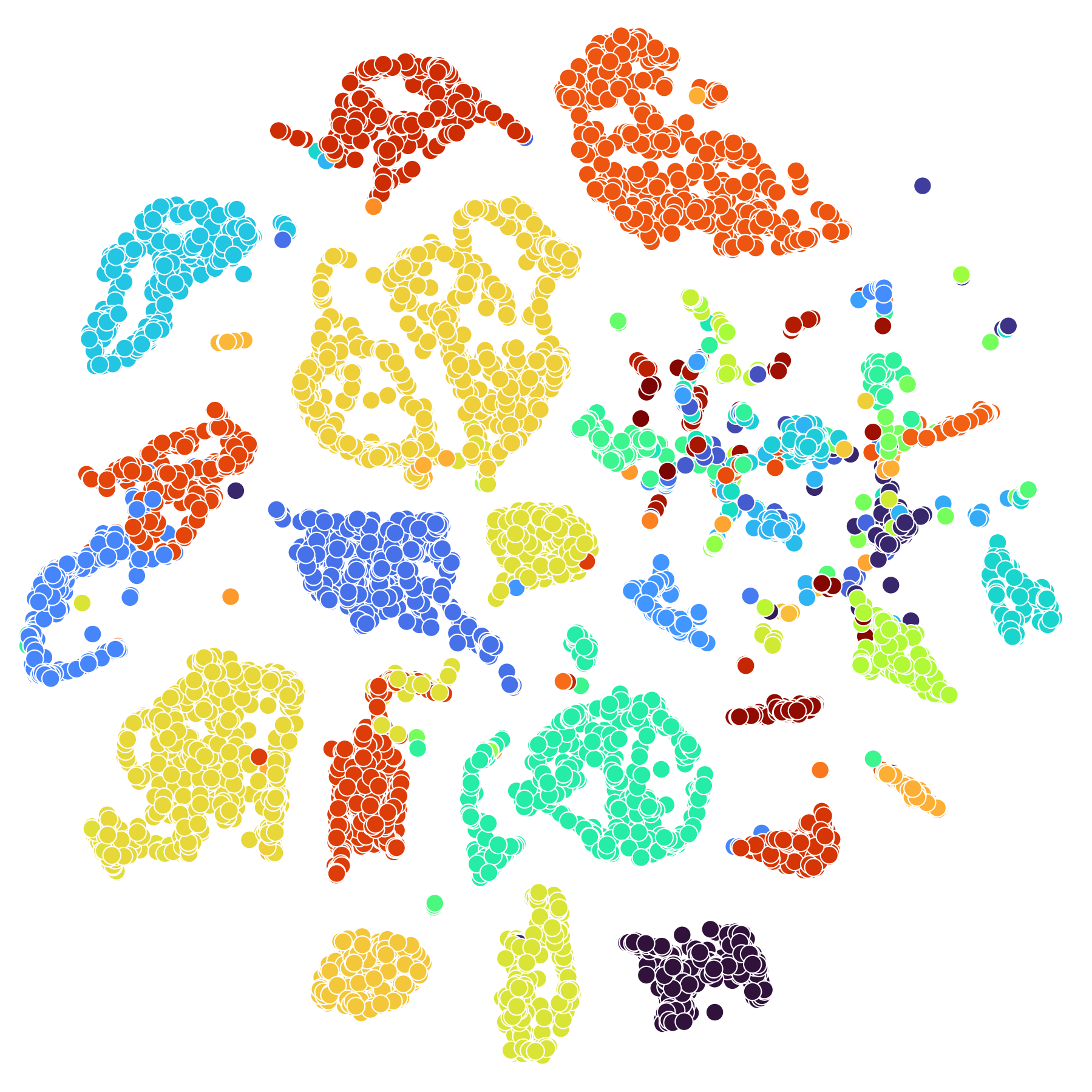}}
\caption{The t-SNET plots of the proposed five models on the MCB-A testing set. Perplexity and iterate are set to 40 and 300, respectively.}
\label{fig4}
\end{figure}
\subsubsection{Retrieval results}
The retrieval results of models are presented in Table \ref{tab5}. Consistent with \cite{he2020improved}, we evaluate the model performance using F1@N, MAP, and NDCG as evaluation indicators. Additionally, we introduce the ``microALL+macroALL" metric, which represents the average performance of microALL and macroALL evaluations, providing a comprehensive assessment. It can be seen that our GMViT achieves the best results in general. Despite slight superiority in the classification task, RotationNet \cite{kanezaki2018rotationnet} does not exhibit superior performance in retrieval. This sensitivity is attributed to the presence of numerous unaligned shapes in MCB-A, which affects RotationNet's performance significantly. PointCNN \cite{li2018pointcnn} and PointNet++ \cite{qi2017pointnet++}, being inherently resistant to point cloud permutation invariance, attain optimal results in \cite{he2020improved}. Furthermore, our small models, GMViT-simple and GMViT-mini, exhibit impressive performance even without knowledge distillation, which is further enhanced through the distillation process. Remarkably, knowledge distillation results in GMViT-simple surpassing GMViT in MAP within the macroALL evaluation. To better observe the similarity of the 3D shape descriptors, we plot them in the Fig. \ref{fig4}. 

\subsection{Analysis of GMViT}
In this section, we analyze the various parameters and components of GMViT. All experiments were carried out under the Dodecahedron-20 setting.

\begin{figure}[t]
	\centering
	\subfloat[ModelNet40]{\includegraphics[width = 0.23\textwidth]{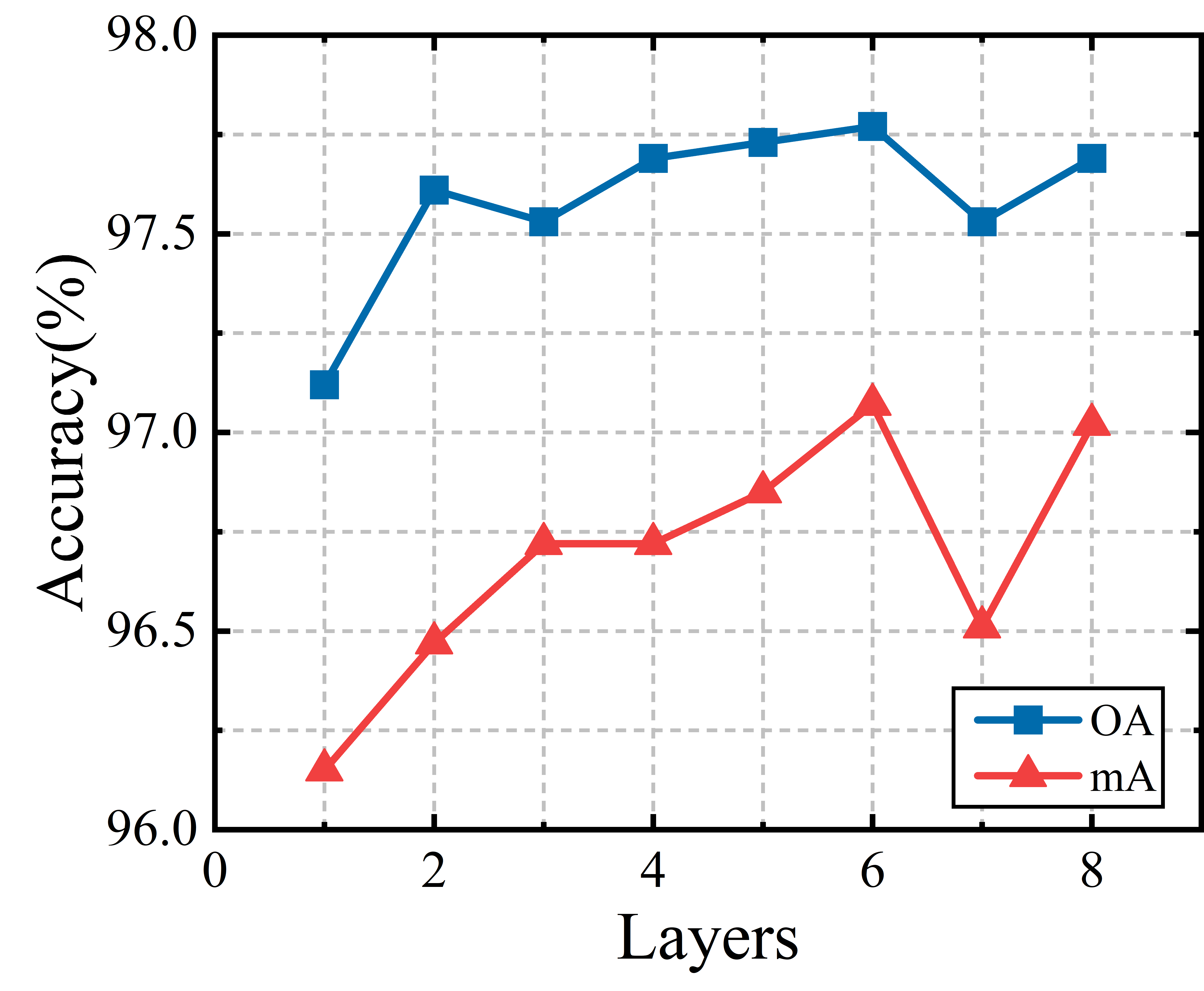}}
	\hfill
	\subfloat[ModelNet10]{\includegraphics[width = 0.23\textwidth]{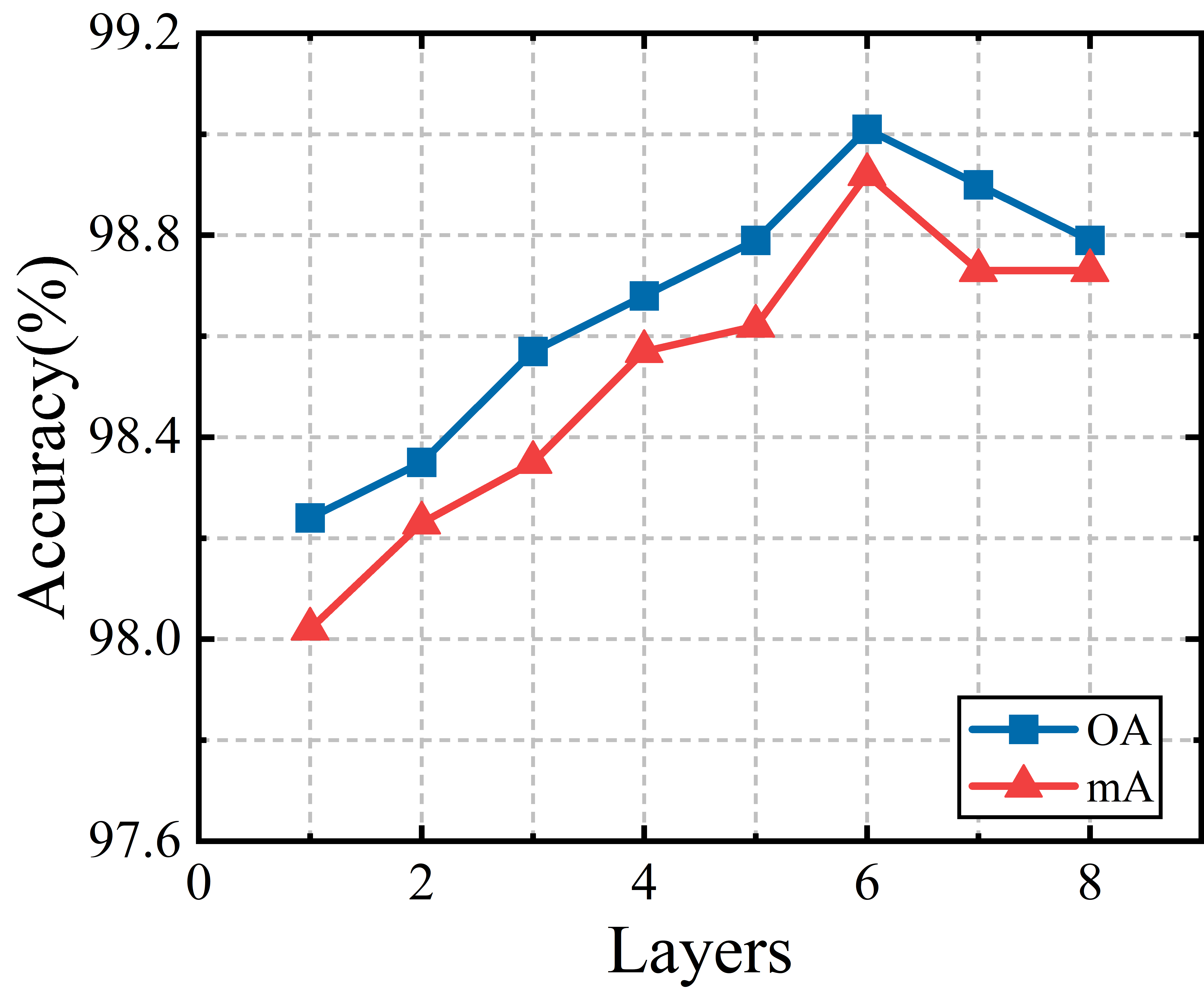}}
	\caption{Classification results of GMViT with different number of ViT layers on ModelNet.}
	\label{fig5}
\end{figure}
\begin{figure}[t]
	\centering
	\subfloat[ModelNet40]{\includegraphics[width = 0.22\textwidth]{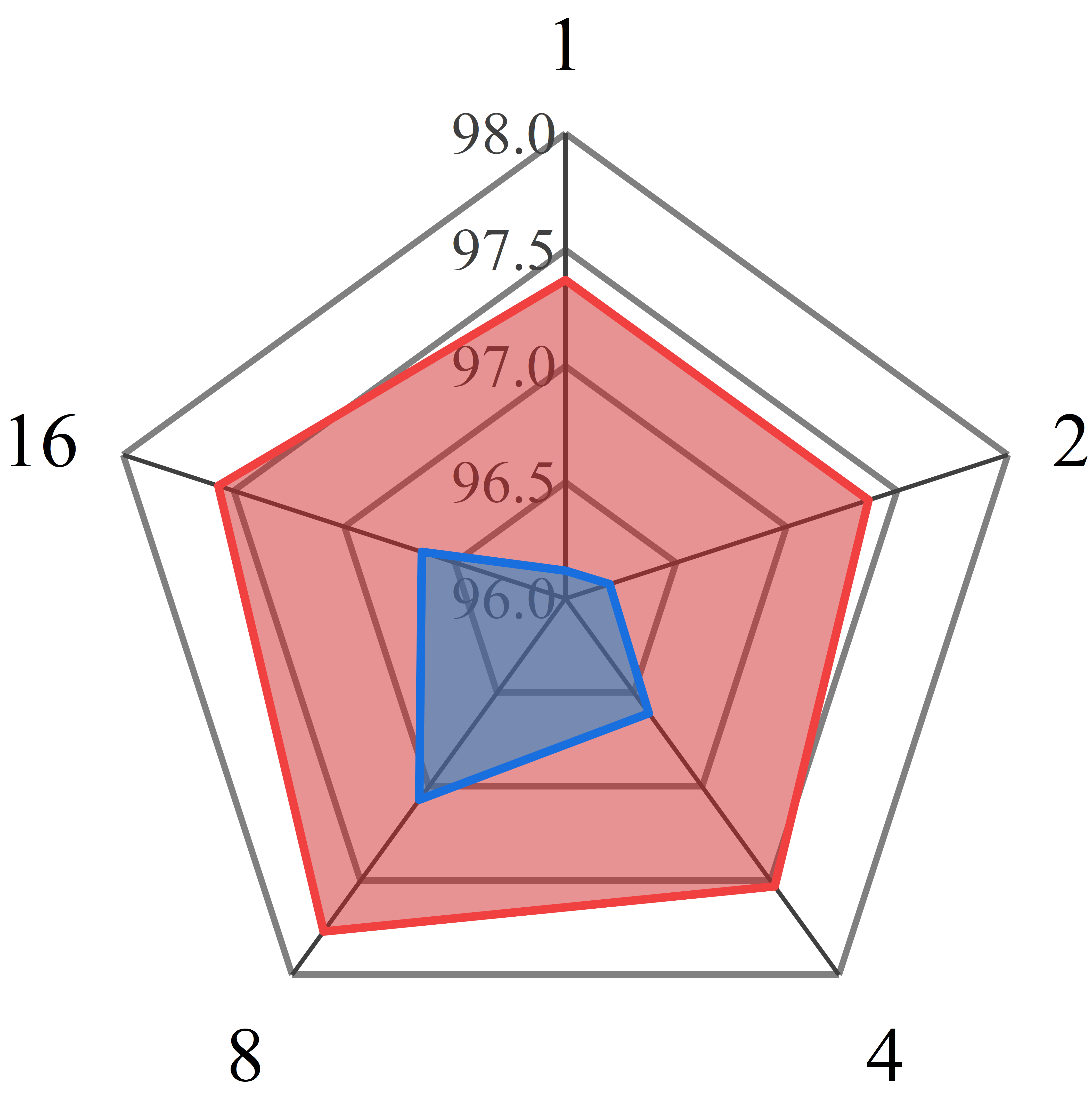}}
	\hfill
	\subfloat[ModelNet10]{\includegraphics[width = 0.22\textwidth]{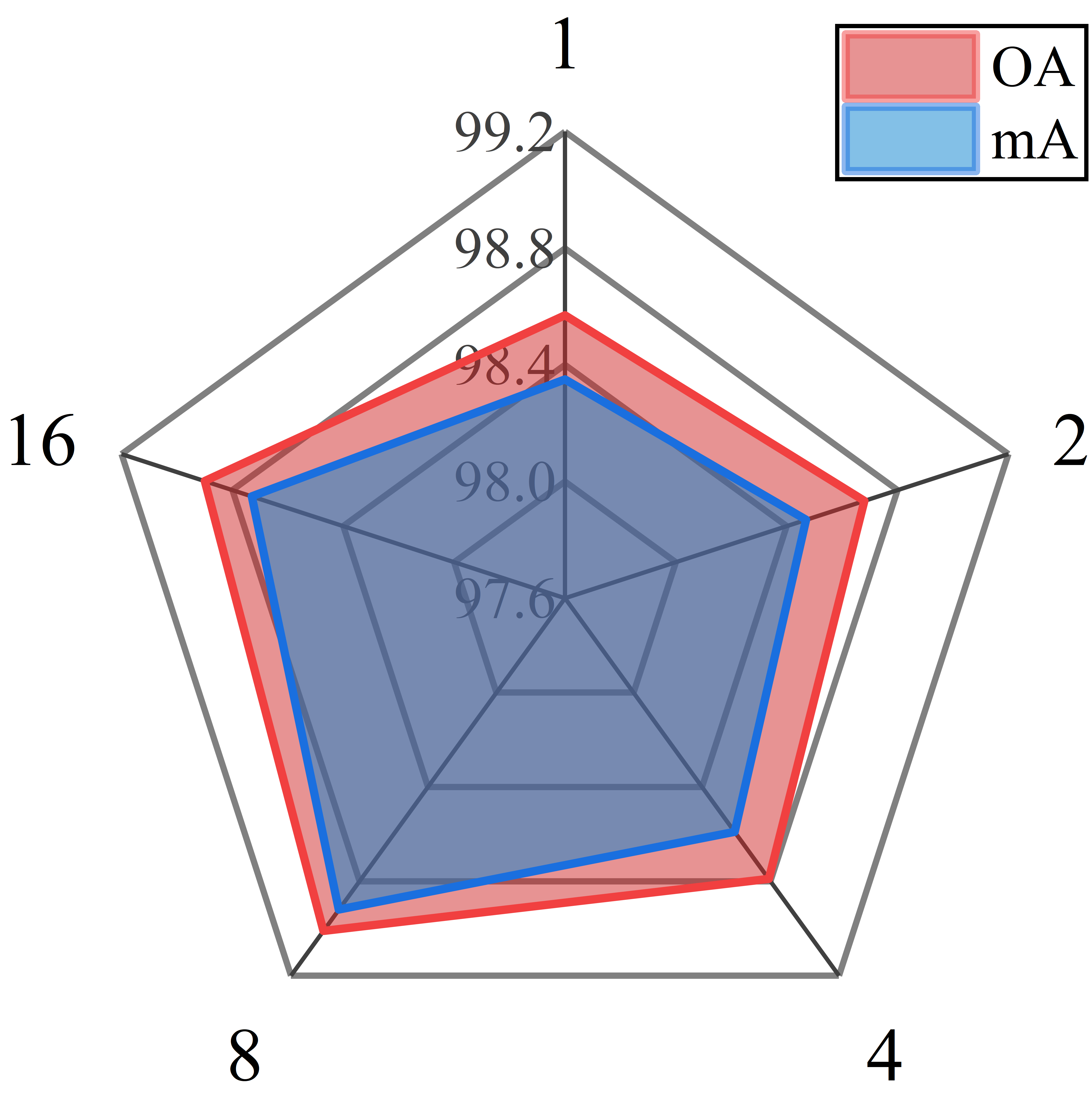}}
	\caption{Classification results of GMViT with different number of self-attention heads on ModelNet.}
	\label{fig6}
\end{figure}
\subsubsection{Position embedding}
We conduct a comparison between the proposed position embedding method and other approaches, and the results are presented in Table \ref{tab6}. Applying the traditional position embedding (PE) improves the OA and mA of the model to some extent compared to the model without PE.  Furthermore, utilizing the camera position as the PE leads to the highest classification performance, improving it by at least 1\% compared to the traditional PE. These findings highlight the significant loss of valuable information when disregarding the positional relationship among views, with the spatial relationship between views containing more crucial information compared to the sequence relationship.

\subsubsection{Number of ViT layers}
We test the classification performance of GMViT by changing the number of stacked layers of ViT. Both the view-level ViT and group-level ViT consist of an equal number of layers. The classification results on ModelNet are presented in Fig. \ref{fig5} (a) and (b). The results demonstrate a consistent increase in accuracy as the number of layers increases from 1 to 6, suggesting that a greater number of layers promotes enhanced interaction between view-level and group-level features. Nevertheless, the accuracy declines with further increases in the number of layers, as the model performance reaches its peak at 6 layers.

\subsubsection{Number of attention heads in ViT} 
We analyze the number of self-attention heads in GMViT. The experimental results showed in Fig. \ref{fig6} (a) demonstrate that the OA of the model falls below 97.5\% on ModelNet40 when using 1 or 2 attention heads, surpasses 97.5\% with 4 attention heads, and achieves its peak performance with 8 attention heads. This indicates that distinct self-attention heads effectively capture diverse semantic information, and the aggregation of multiple heads enriches the final 3D representation. Nonetheless, setting the number of heads to 16 leads to a decrease in model accuracy, possibly due to information redundancy arising from an excessive number of attention points. Fig. \ref{fig6} (b) also demonstrates the same accuracy trend of the model on ModelNet10.

\begin{figure}[t]
	\centering
	\subfloat[ModelNet40]{\includegraphics[width = 0.22\textwidth]{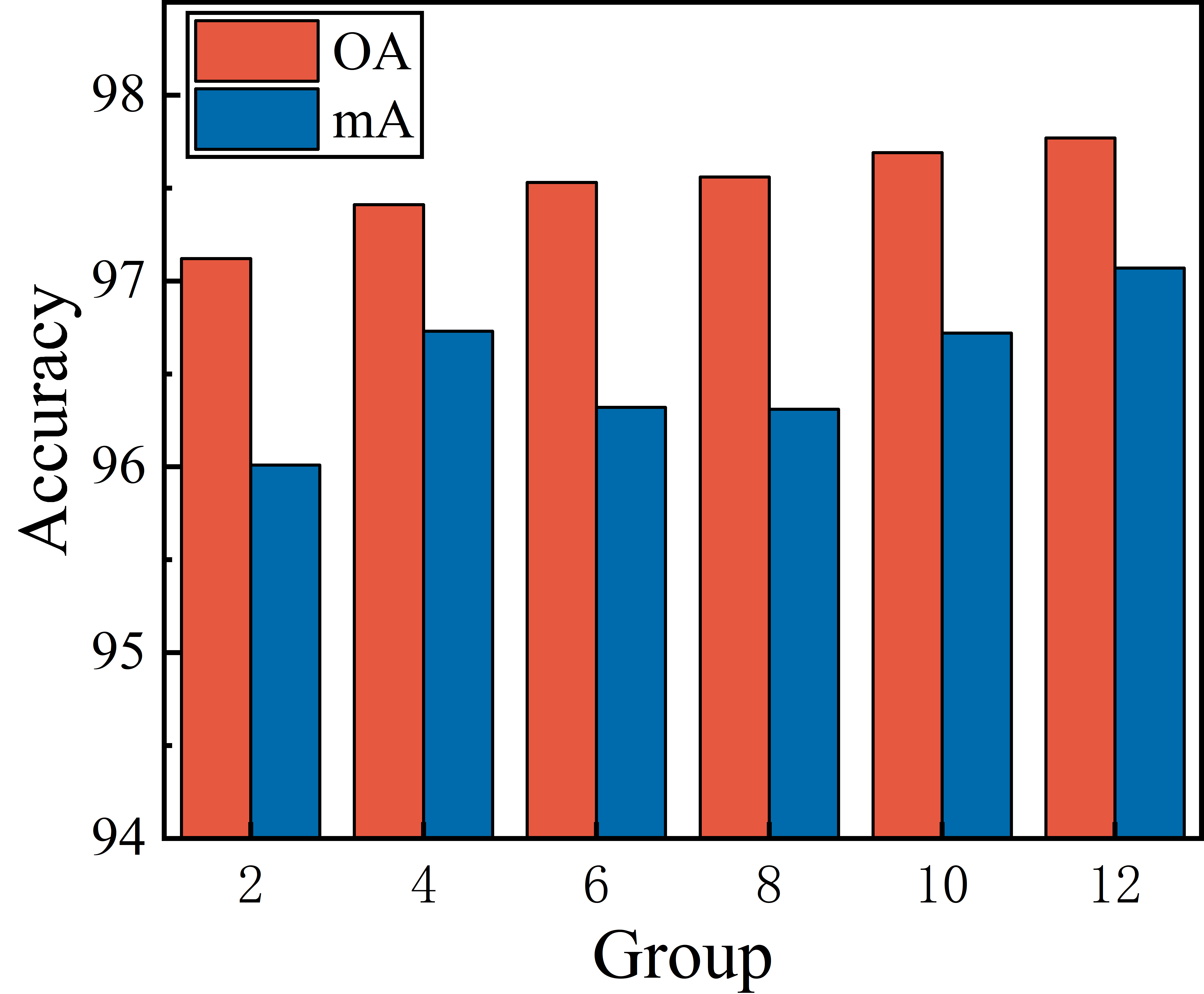}}
	\hfill
	\subfloat[ModelNet10]{\includegraphics[width = 0.23\textwidth]{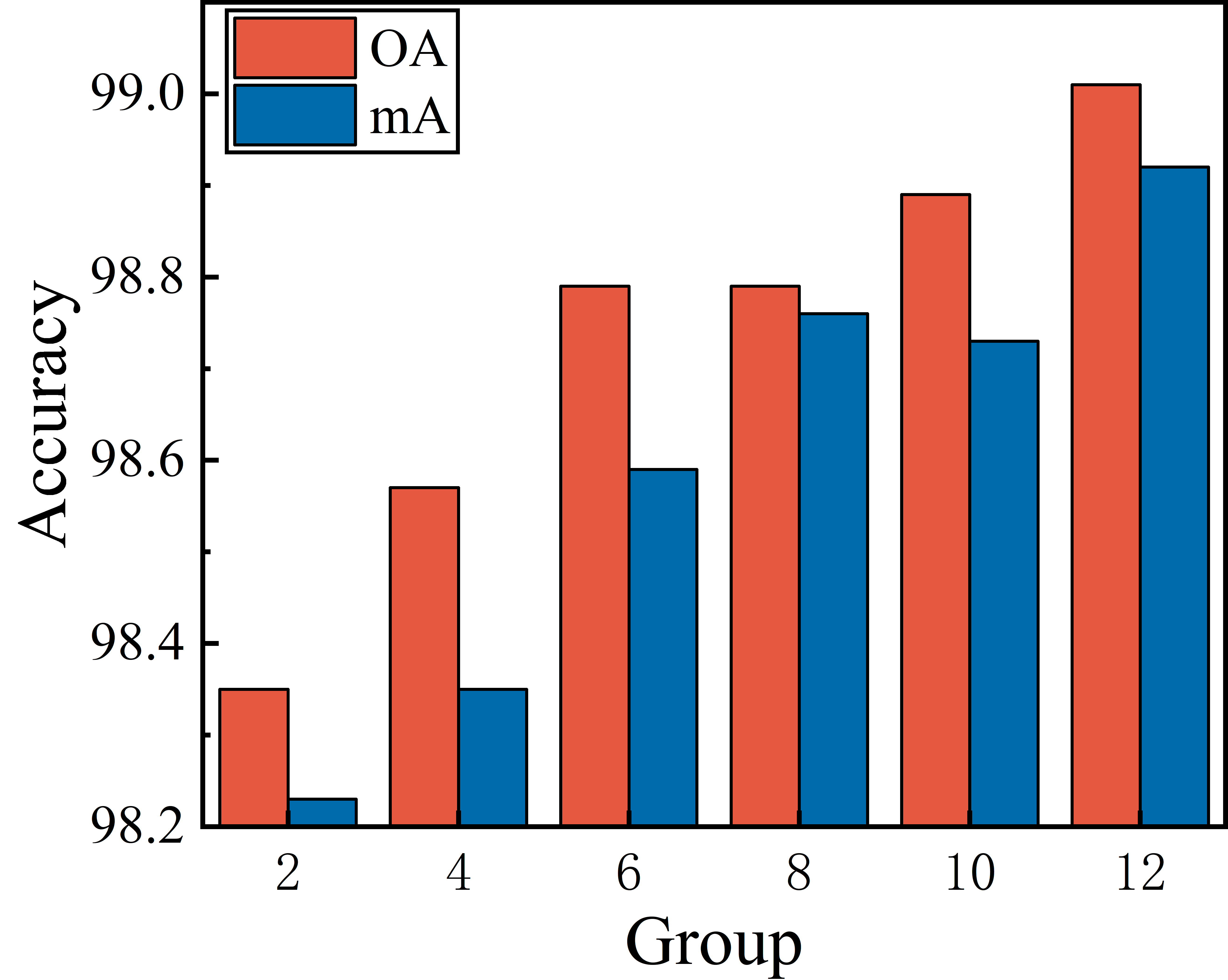}}
	\caption{Classification results of GMViT with different number of groups on ModelNet.}
	\label{fig7}
\end{figure}
\begin{table}[t]
	\caption{Classification results of GMViT with different position embedding (PE) on ModelNet40.}
	\centering
	\setlength{\tabcolsep}{8pt}
	\begin{tabular}{|l|cc|}
		\hline
		Model                  & OA(\%)         & mA(\%)         \\ \hline
		GMViT(without PE)      & 96.31          & 94.72          \\
		GMViT(conventional PE) & 96.68          & 95.55          \\
		GMViT                  & \textbf{97.77} & \textbf{97.07} \\ \hline
	\end{tabular}
	\label{tab6}
\end{table}
\begin{table}[!t]
	\centering
	\caption{Classification results of GMViT fitted with different components on ModelNet40.}
	\setlength{\tabcolsep}{1mm}{
		\begin{tabular}{|ccc|cc|}
			\hline
			View-level ViT & Group module & Group-level ViT & OA(\%) & mA(\%) \\ \hline
			\checkmark              &              &                  & 97.33  & 96.29  \\
			\checkmark              &              & \checkmark                & 97.20   & 96.31  \\
			& \checkmark            & \checkmark                & 96.80   & 95.46  \\
			\checkmark              & \checkmark            & \checkmark                & \textbf{97.77}  & \textbf{97.07}  \\ \hline
	\end{tabular}}
	\label{tab7}
\end{table}
\subsubsection{Number of groups of grouping modules}
We observe the change of GMViT on ModelNet by changing the number of groups of GMViT grouping modules. The Fig. \ref{fig7} (a) and (b) illustrate consistent increases in overall accuracy (OA) as the number of groups increases from 2 to 12, demonstrating that finer groupings enhance the model's performance. While it is not guaranteed that each group can be assigned features among the numerous divisions, a larger number of groups refines the boundaries of each group. Across various models employing different grouping modules, objects with the same view group token may yield different groupings due to variations in the degree of group boundaries.

\subsubsection{Components of GMViT}
Finally we conduct ablation analysis on the view-level ViT, grouping module and group-level ViT of GMViT. The classification results of various GMViT versions on ModelNet40 are presented in Table \ref{tab7}. In the absence of a grouping module, group-level features are nonexistent, making the group-level ViT equivalent to the view-level ViT. Consequently, models with fewer layers of ViT outperform those with more layers in terms of performance. The absence of the view-level ViT has the most detrimental impact on the model's classification performance. This could be attributed to the lack of information interaction between the view features generated by the CNN, as they are directly grouped and pooled within the grouping module, leading to significant information loss. This confirms the indispensable role of all three components in GMViT.

\begin{table}[t]
\centering
\caption{Comprehensive comparison of various methods on ModelNet40 dataset. The bold values represent the parameter compression multiplier, distillation performance preservation rate and inference speed multiplier of the small model, respectively.}
\setlength{\tabcolsep}{0.5mm}
\begin{tabular}{|c|c|c|c|c|c|}
\hline
Views               & Model            & \begin{tabular}[c]{@{}c@{}}\#Param.\\ (M)\end{tabular}                 & \begin{tabular}[c]{@{}c@{}}Classification\\ OA(\%)\end{tabular} & \begin{tabular}[c]{@{}c@{}}Retrieval\\ mAP(\%)\end{tabular} & \begin{tabular}[c]{@{}c@{}}Inference\\ speed\end{tabular}               \\ \hline
\multirow{13}{*}{12} & MVCNN \cite{su2015multi}           & 128.9                                                                  & 89.5                                                            & 80.2                                                        & 24.3                                                                    \\
                    & GVCNN \cite{feng2018gvcnn}           & 41.2                                                                   & 92.6                                                            & 85.7                                                        & 17.5                                                                    \\
                    & MVDAN \cite{wang2022multi}           & 23.7                                                                   & 96.6                                                            & -                                                           & 31.1                                                                   \\ 
                    \cline{2-6} 
                    & GMViT            & 44.1                                                                   & 96.27                                                           & 94.54                                                       & 55.1                                                                    \\ \cline{2-6} 
                    & GMViT-simple     & \multirow{2}{*}{\begin{tabular}[c]{@{}c@{}}5.5\\ \scriptsize{(\textbf{8×})}\end{tabular}}    & 91.9                                                            & 86.19                                                       & \multirow{2}{*}{\begin{tabular}[c]{@{}c@{}}79.7\\ \scriptsize{(\textbf{1.45×})}\end{tabular}} \\
                    & GMViT-simple(KD) &                                                                        & 92.95\scriptsize{(\textbf{96.6\%})}                                                   & 90.54\scriptsize{(\textbf{95.8\%})}                                               &                                                                         \\ \cline{2-6} 
                    & GMViT-mini       & \multirow{2}{*}{\begin{tabular}[c]{@{}c@{}}2.5\\ \scriptsize{(\textbf{17.6×})}\end{tabular}} & 89.55                                                           & 80.88                                                       & \multirow{2}{*}{\begin{tabular}[c]{@{}c@{}}91.4\\ \scriptsize{(\textbf{1.66×})}\end{tabular}} \\
                    & GMViT-mini(KD)   &                                                                        & 92.42\scriptsize{(\textbf{96\%})}                                                     & 85.84\scriptsize{(\textbf{90.8\%})}                                               &                                                                         \\ \hline
\multirow{10}{*}{20} & View-GCN \cite{wei2020view}        & 33.9                                                                   & 97.6                                                            & -                                                           & 39.8                                                                    \\
                    & RotationNet \cite{kanezaki2018rotationnet}     & 24.2                                                                   & 97.37                                                           & -                                                           & 23.1                                                                    \\ 
                    \cline{2-6} 
                    & GMViT            & 44.1                                                                   & 97.77                                                           & 97.57                                                       & 33.0                                                                      \\
                    \cline{2-6} 
                    & GMViT-simple     & \multirow{2}{*}{\begin{tabular}[c]{@{}c@{}}5.5\\ \scriptsize{(\textbf{8×})}\end{tabular}}    & 95.06                                                           & 89.44                                                       & \multirow{2}{*}{\begin{tabular}[c]{@{}c@{}}41.1\\ \scriptsize{(\textbf{1.25×})}\end{tabular}} \\
                    & GMViT-simple(KD) &                                                                        & 95.75\scriptsize{(\textbf{97.9\%})}                                                   & 94.24\scriptsize{(\textbf{96.6\%})}                                               &                                                                         \\ \cline{2-6} 
                    & GMViT-mini       & \multirow{2}{*}{\begin{tabular}[c]{@{}c@{}}2.5\\ \scriptsize{(\textbf{17.6×})}\end{tabular}} & 93.44                                                           & 87.36                                                       & \multirow{2}{*}{\begin{tabular}[c]{@{}c@{}}47.5\\ \scriptsize{(\textbf{1.44×})}\end{tabular}} \\
                    & GMViT-mini(KD)   &                                                                        & 95.75\scriptsize{(\textbf{97.9\%})}                                                   & 91.12\scriptsize{(\textbf{93.4\%})}                                               &                                                                         \\ \hline
\end{tabular}
\label{tab8}
\end{table}
\subsection{Analysis of knowledge distillation}
\subsubsection{Compression effect}
\begin{figure}[t]
	\centering
	\includegraphics[width=0.9\linewidth]{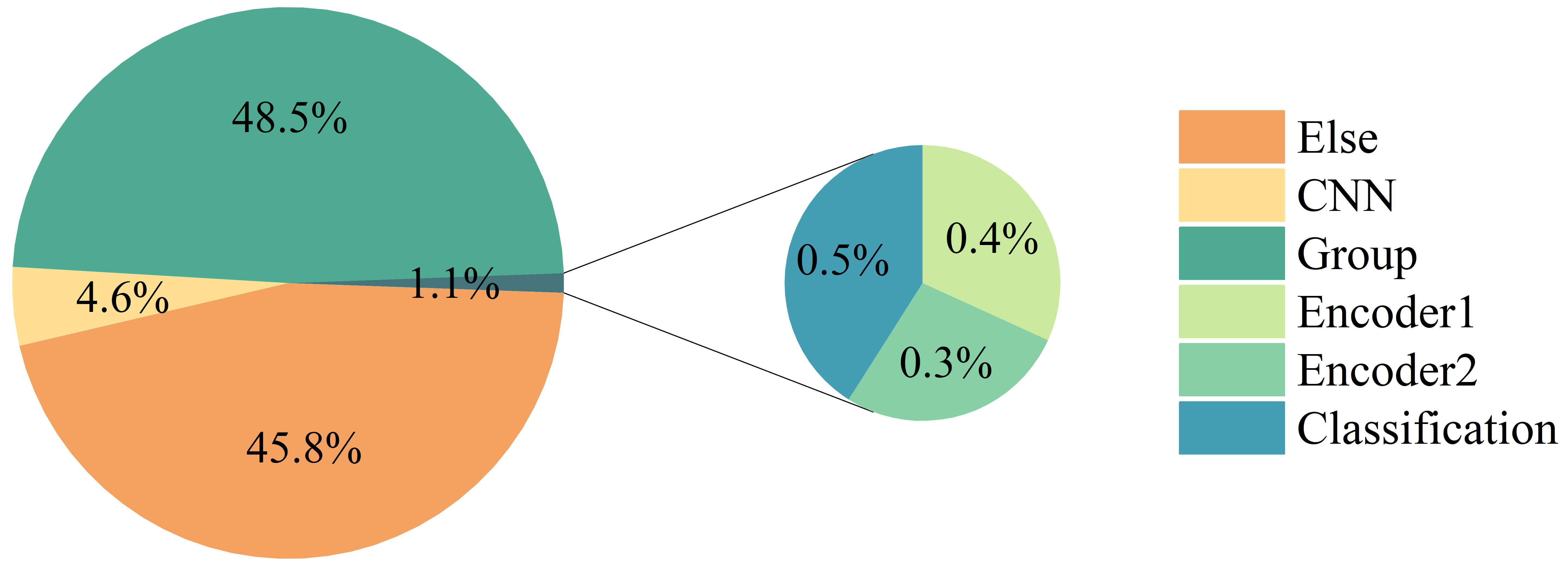}
	\caption{GMViT-mini time spent by modules within one epoch on the ModelNet40 testing set.}
	\label{fig8}
\end{figure}

We analyze the impact of model compression. To ensure a fair comparison, all models are tested on a single NVIDIA RTX 3090 GPU. BatchSize is set to 8 to account for system memory variations, and experiments are conducted on the ModelNet40 testing set. Inference speed is measured in objects per second, calculated by the time taken for the model to classify objects within a single epoch. Results are shown in Table \ref{tab8}. Despite MVCNN's larger VGG-M baseline and the highest parameter count, it achieves the lowest classification and retrieval results. GMViT, although not having the largest parameter size, outperforms most methods in both inference speed and performance. Notably, GMViT-simple and GMViT-mini are compressed versions of GMViT, reducing parameter size by 8 and 17.6 times, respectively, while maintaining at least 96\% and 90\% of the classification and retrieval performance through knowledge distillation. Our small model exhibits approximately 1.5 times faster inference speed compared to the large model, a modest improvement considering the significant reduction in parameter size. The time distribution analysis in Fig. \ref{fig8} reveals that the majority of processing time is allocated to the "Group" and "Else" components, likely due to the dominance of looping statements in these sections. The limited increase in inference speed can be attributed to the shared use of the same grouping modules in both the large and small models.

\begin{table}[t]
	\centering
	\scriptsize
	\caption{Ablation analysis of different distillation targets on ModelNet40.}
	\setlength{\tabcolsep}{0.9mm}{
		\begin{tabular}{|cc|c|c|ccc|cc|}
			\hline
			\multicolumn{2}{|c|}{logit} & \begin{tabular}[c]{@{}c@{}}global\\ feature\end{tabular} & \begin{tabular}[c]{@{}c@{}}group \\ token\end{tabular} & \multicolumn{3}{c|}{\begin{tabular}[c]{@{}c@{}}intermediate\\ features\end{tabular}} & \multirow{2}{*}{OA(\%)} & \multirow{2}{*}{mA(\%)} \\ \cline{1-7}
			$\mathcal{L}_{hard}$            & $\mathcal{L}_{soft}$            & $\mathcal{L}_{global}$                                                        & $\mathcal{L}_{token}$                                                      & $\mathcal{L}_{group}$                          & $\mathcal{L}_{view}$                          & $\mathcal{L}_{CNN}$                          &                         &                         \\ \hline
			\checkmark            &              &                                                          &                                                        &                            &                            &                            & 88.70                    & 86.01                   \\
			\checkmark            & \checkmark            &                                                          &                                                        &                            &                            &                            & 89.63                   & 86.37                   \\
			\checkmark            & \checkmark            & \checkmark                                                        &                                                        &                            &                            &                            & 90.32                   & 86.62                   \\
			\checkmark            & \checkmark            & \checkmark                                                        & \checkmark                                                      &                            &                            &                            & 90.48                   & 86.78                   \\
			\checkmark            & \checkmark            & \checkmark                                                        & \checkmark                                                      & \checkmark                          &                            &                            & 91.53                   & 88.19                   \\
			\checkmark            & \checkmark            & \checkmark                                                        & \checkmark                                                      & \checkmark                          & \checkmark                          &                            & 92.18                   & 88.60                    \\
			\checkmark            & \checkmark            & \checkmark                                                        & \checkmark                                                      & \checkmark                          & \checkmark                          & \checkmark                          & \textbf{92.42}          & \textbf{88.99}          \\ \hline
	\end{tabular}}
	\label{tab11}
\end{table}
\subsubsection{Distillation targets}
The inclusion or exclusion of distillation losses directly signifies the presence or absence of the distillation targets. As shown in Table \ref{tab11}, the model's performance keeps improving as we gradually increase the distillation target. The incremental improvements validate the rationale behind each distillation target: CNN features provide basic view representations to transfer lower-level knowledge. View-level ViT outputs contain complex relational information that is difficult to learn alone, providing sophisticated feature distillation. Group-level ViT outputs further enrich the relational information transfer. Group tokens transfer crucial grouping knowledge, demonstrating the value of distilling information-rich intermediate outputs. Global features provide holistic supervision, in line with distilling the most influential outputs. Logit distillation gives end-to-end guidance. Additionally, the targets cover both low-level view features and high-level shape representations, enabling multi-scale knowledge transfer. The substantial improvements from view-level ViT group-level align with the strategy of distilling complex, information-rich module outputs. The global feature improvements validate distilling influential intermediate results. In conclusion, the multi-faceted targets effectively transfer knowledge at different levels of abstraction and complexity, leading to optimized student learning. The analysis demonstrates principled distillation target selection.

\subsubsection{Distillation temperature}
The impact of various temperatures on the distillation effect is presented in Table \ref{tab9}. At a temperature value of 1, the student model achieves an OA of only 88.53\%, which is inferior to the performance of the model trained without distillation. This suggests that at this temperature, the soft labels entirely preserve the teacher model's output, making it challenging for the student model to learn the complex details. In contrast, the classification performance of the student model reaches its peak when the temperature is raised to 5, implying that higher temperatures facilitate the student model's learning from the teacher model. Nevertheless, as the temperature further increases, the performance of the student model deteriorates, possibly attributable to the over-smoothing of the soft labels caused by the excessively high temperature.

\begin{table}[t]
\centering
\caption{Different distillation temperatures(T) on ModelNet40.}
\setlength{\tabcolsep}{8pt}
\begin{tabular}{|c|cc|}
\hline
Temperature & OA(\%)         & mA(\%)         \\ \hline
1           & 88.53          & 85.56          \\
2           & 90.52          & 87.05          \\
3           & 91.49          & 87.74          \\
4           & 91.65          & 88.24          \\
5           & \textbf{92.42} & \textbf{88.99} \\
6           & 91.69          & 88.25          \\
7           & 91.90          & 88.73          \\
8           & 92.18          & 88.82          \\
9           & 91.82          & 88.57          \\
10          & 92.06          & 88.60           \\
11          & 91.53          & 87.96          \\
12          & 91.69          & 88.65          \\ \hline
\end{tabular}
\label{tab9}
\end{table}
\begin{table}[h]
\centering
\caption{Different soft and hard label coefficients on ModelNet40.}
\begin{tabular}{|cc|cc|}
\hline
$\mathcal{L}_{soft}$ & $\mathcal{L}_{hard}$ & OA(\%) & mA(\%) \\ \hline
0.0  & 1.0    & 92.06  & 88.43  \\
0.1  & 0.9  & 92.18  & 88.96  \\
0.2  & 0.8  & 91.33  & 87.84  \\
0.3  & 0.7  & 91.82  & 88.56  \\
0.4  & 0.6  & 91.98  & 88.67  \\
0.5  & 0.5  & 92.01  & 88.87  \\
0.6  & 0.4  & 92.22  & 88.47  \\
0.7  & 0.3  & \textbf{92.42}  & \textbf{88.99}  \\
0.8  & 0.2  & 92.10   & 88.93  \\
0.9  & 0.1  & 92.26  & 88.30   \\
1.0    & 0.0    & 92.10  & 88.89  \\ \hline
\end{tabular}
\label{tab10}
\end{table}
\subsubsection{Coefficients of soft and hard labels}
To enhance the training of the student model, we perform experiments to determine the optimal label coefficients. As shown in Table \ref{tab10}, we vary the coefficients of the soft label and hard label from 0 to 1 while ensuring their sum is 1. The model attains optimal classification results with coefficients of 0.7 for the soft label and 0.3 for the hard label. Setting the hard label coefficient to 0 leads to a degradation in the model's classification performance, suggesting that the teacher model's conclusions are not always reliable during the student model's learning process, and the hard label is necessary to rectify errors when required. Similarly, when the soft label coefficient is set to 0, the model's performance is diminished, indicating that the soft label encompasses more meaningful information than the hard label.

\section{Conclusion}
\label{sec5}
In this paper, we propose a method called Group Multi-view Vision Transformer (GMViT) for 3D shape recognition. To strengthen view relationships, we utilize view-level ViT to foster interaction among view-level features. For capturing information at varying scales, we employ the grouping module to aggregate low-level view-level features into high-level group-level features. Additionally, we employ group-level ViT to fuse the group-level features and obtain the final 3D shape descriptor. Notably, The introduced spatial encoding of camera coordinates as position embeddings equips the model with valuable view spatial information. GMViT has exhibited outstanding performance on multiple 3D shape recognition datasets. 

Furthermore, we pioneer application of knowledge distillation to multi-view 3D shape recognition, enabling model compression while preserving performance. The distillation incorporates complementary outputs to transfer multi-scale knowledge. This systematic approach effectively transfers knowledge across different levels of abstraction, as demonstrated by substantial improvements. While promising, some limitations exist in  distillation speed-up. Future work can address this and extend the method to other 3D tasks.


\end{document}